\documentclass{article}

\PassOptionsToPackage{square,comma,numbers,sort&compress}{natbib}

\usepackage[main,final]{neurips_2026}

\usepackage[utf8]{inputenc} %
\usepackage[T1]{fontenc}    %
\usepackage[table,dvipsnames]{xcolor}
\definecolor{cvprblue}{rgb}{0.21,0.49,0.74}
\usepackage[breaklinks,colorlinks,citecolor=cvprblue]{hyperref}
\usepackage{url}            %
\usepackage{booktabs}       %
\usepackage{amsfonts}       %
\usepackage{nicefrac}       %
\usepackage{microtype}      %
\usepackage{amsmath}
\usepackage{graphicx}
\usepackage{multirow}
\usepackage{caption}
\usepackage{enumitem}
\usepackage{wrapfig}
\usepackage{algorithm}
\usepackage{algpseudocode}
\usepackage{placeins}
\usepackage{tabularx}
\usepackage[most]{tcolorbox}
\usepackage{array}
\definecolor{recaGreen}{RGB}{244,252,246}
\definecolor{privGray}{RGB}{130,130,130}
\newtcolorbox[auto counter, number within=section]{promptbox}[2][]{
    breakable,
    enhanced,
    width=\linewidth,
    colback=gray!4,
    colframe=black!55,
    fonttitle=\bfseries,
    boxrule=0.6pt,
    arc=2pt,
    left=5pt,
    right=5pt,
    top=5pt,
    bottom=5pt,
    title={Box~\thetcbcounter: #2},
    label={#1}
}

\title{Drag as Evidence: Motion-Grounded Latent Recomposition for Drag-Based Editing}

\author{%
  Xinyu Pu$^{1,3}$ \qquad
  Hongsong Wang$^{2}$\thanks{Corresponding authors.} \qquad
  Jie Gui$^{1,3}$\footnotemark[1] \qquad
  Pan Zhou$^{4}$ \\
  $^{1}$School of Cyber Science and Engineering, Southeast University, Nanjing, China \\
  $^{2}$School of Computer Science and Engineering, Southeast University, Nanjing, China \\
  $^{3}$Purple Mountain Laboratories, Nanjing, China \\
  $^{4}$Singapore Management University, Singapore \\
  \texttt{\{xinyupu, hongsongwang, guijie\}@seu.edu.cn}, \texttt{panzhou3@gmail.com}
}

\begin{document}

\maketitle

\begin{abstract}
Modern image editors excel at semantic manipulation and visual synthesis, yet remain limited in precise spatial control, motivating the development of drag-based editing. However, existing drag-based methods often struggle to balance drag accuracy with natural, plausible, and intent-aligned generation. We propose \emph{MoRe-Drag}, a motion-grounded drag-based editing method. Our key insight is to treat pixel-space warping as coarse motion evidence, and to inject this evidence into the generative sampling trajectory. Specifically, \emph{MoRe-Drag} performs region-aware latent recomposition over refinement, inpainting, and preservation regions, coupled with stage-adaptive conditioning that progressively shifts from motion-grounded structure formation to semantic refinement. We further support a no-manual-instruction interface by adapting the MLLM-based text encoder for drag-aware instruction inference. Experiments on \textsc{DragBench-SR} and \textsc{DragBench-DR} show that \emph{MoRe-Drag} substantially improves drag precision over strong base editors and achieves superior drag accuracy among SOTA drag-based methods, while delivering strong semantic consistency and visually realistic results. Project page: \url{https://xinyu-pu.github.io/projects/more-drag/}.
\end{abstract}

\begin{figure}[h]
    \centering
    \includegraphics[width=1\textwidth]{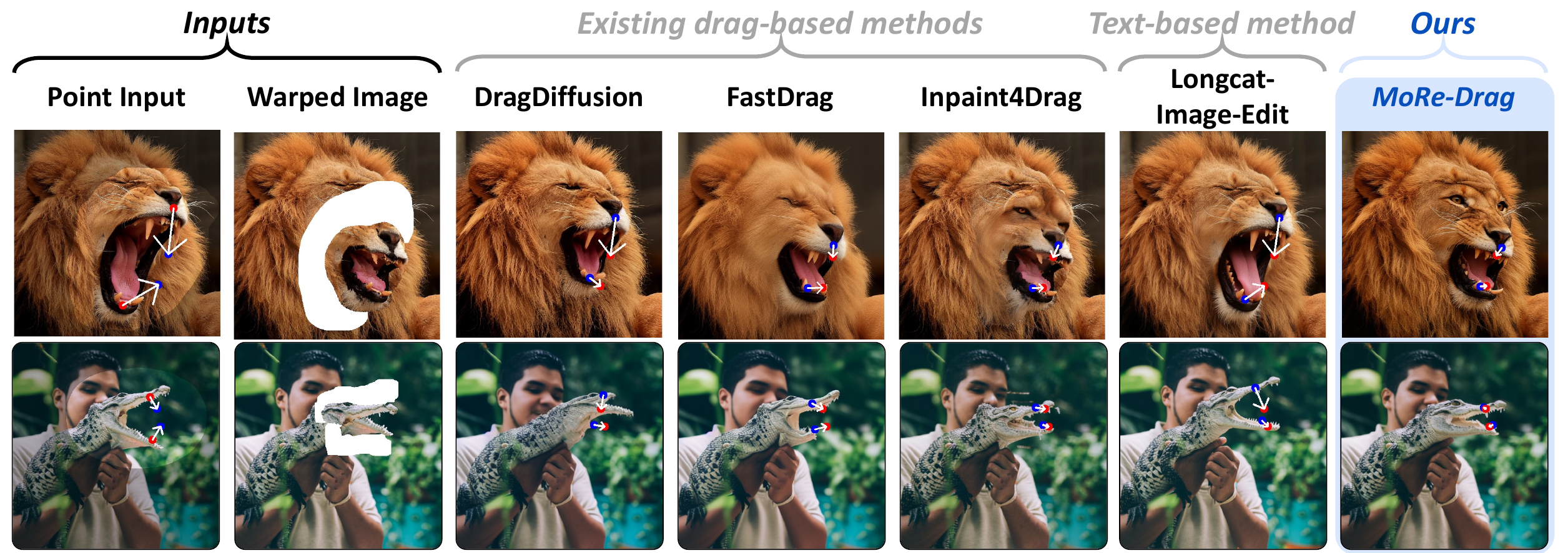}
    \caption{\textbf{Editing comparison.} Pixel-space warping provides explicit but imperfect motion evidence. \emph{MoRe-Drag} injects this evidence into the generative trajectory, enabling the editor to follow the specified motion while restoring natural structure and appearance.
    }
    \label{fig:teaser}
\end{figure}

\section{Introduction}
\begin{figure}
    \centering
    \includegraphics[width=\linewidth]{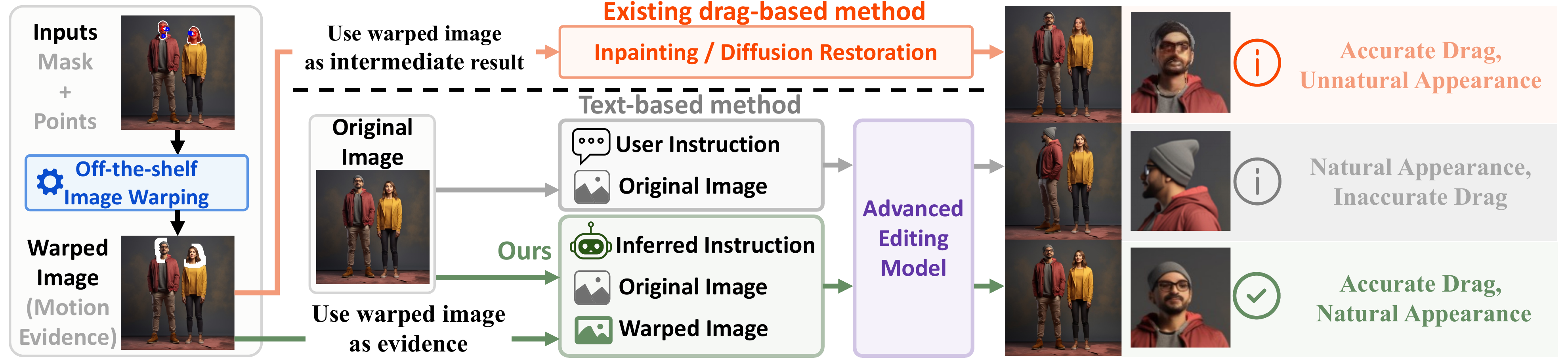}
    \caption{\textbf{Conceptual comparison with existing drag-based and text-based editing paradigms.} Existing drag methods use warped image as an intermediate result, leading to accurate but unnatural edits, while text-based editors produce natural results without precise drag control. Our method uses warped image as motion evidence to achieve both accurate dragging and natural appearance.}
    \label{fig:simple_compare}
\end{figure}

Image generative models have advanced rapidly in realism, instruction following, and identity preservation~\cite{DBLP:conf/iclr/ChoLKOJ24,DBLP:journals/corr/abs-2503-19839,hertz2022p2p,DBLP:conf/icml/0006YMXE024,zhang2025ICEdit,DBLP:journals/corr/abs-2506-18871,DBLP:journals/ijautcomp/LuZBCLZ25,DBLP:journals/ijautcomp/ZhangWHWWZZLL25,yan2026morphsig,yan2026animeagent}, driven by progress in data curation~\cite{DBLP:conf/cvpr/YuCYPWW0TZZ25,DBLP:conf/iclr/WeiXRDZC25}, model training~\cite{DBLP:journals/corr/abs-2505-07818,liu2025flow,DBLP:conf/iclr/LipmanCBNL23,DBLP:conf/iclr/LiuG023,DBLP:journals/corr/abs-2605-12112,tan2026gettingspacerightagentic,DBLP:journals/corr/abs-2602-03425}, and architectural improvements~\cite{DBLP:conf/iccv/PeeblesX23,DBLP:journals/corr/abs-2506-15742}. These advances have established natural language as an effective interface for semantic image generation and editing. Recent general-purpose image editing models~\cite{wu2025qwenimagetechnicalreport,DBLP:journals/corr/abs-2512-07584,liu2025step1x-edit} further strengthen this paradigm by leveraging MLLM-based text encoders and joint image-text conditioning, enabling image-aware instruction representations and supporting high-fidelity and visually consistent edits. However, as shown in the gray-shaded middle panel of Fig.~\ref{fig:simple_compare}, text-based instructions are primarily semantic: while effective at specifying what should be changed, they often fall short for spatially grounded manipulations, which require users to specify where the content should be adjusted and to what extent. This limitation motivates drag-based editing as a dedicated interface that provides explicit motion control through handle-to-target point pairs~\cite{draggan,dragdiff}.

\textbf{Motivation.} Existing drag-based editing methods usually preserve the identity and appearance of the input image via image-specific reconstruction, such as LoRA fine-tuning, inversion, or cached key-value features~\cite{dragdiff,fastdrag,regiondrag,DBLP:conf/icml/Xia0SL25}. They then realize drag control either by optimizing latent representations with motion supervision and point tracking~\cite{dragdiff,noisedrag,freedrag,gooddrag,adaptivedrag,DBLP:conf/icml/Xia0SL25,DBLP:conf/iclr/JiangWC25}, or by applying explicit deformation in the latent or pixel space~\cite{fastdrag,regiondrag,DBLP:conf/icml/ShiLYTF25}. Nevertheless, as shown in the top panel of Fig.~\ref{fig:simple_compare}, accurate dragging and natural synthesis remain difficult to reconcile: strong motion constraints can distort object structure, compromise texture consistency, or introduce unnatural artifacts, as also illustrated by existing drag-based methods in Fig.~\ref{fig:teaser}. Conversely, stronger generative regularization can produce realistic images, but often fails to fully satisfy the requested deformation, as seen in the text-based method in Fig.~\ref{fig:teaser}. This precision–naturalness trade-off remains a central obstacle to reliable drag-based editing. This limitation motivates us to revisit drag-based editing through modern general-purpose image editors, whose strong generative priors support realistic and visually consistent synthesis. However, drag interaction requires explicit motion control beyond text-based instruction. The key challenge is therefore \textbf{\textit{how to couple the strong generative prior of modern editors with explicit motion evidence to obtain natural and motion-consistent edits.}}

\textbf{Contribution.} To address this challenge, we make a crucial shift in how pixel-level dragging contributes to the editing process. As illustrated in the bottom green panel of Fig.~\ref{fig:simple_compare}, the warped image is not treated as an intermediate edit, but as {motion evidence} that coarsely encodes the desired deformation. This evidence complements the generative prior of advanced image editors with explicit drag control. Based on this insight, we propose \emph{MoRe-Drag}, a \textbf{Mo}tion-grounded latent \textbf{Re}composition framework, which consists of three key components: region-aware recomposition for motion-evidence, disoccluded, and preservation regions; generation-stage adaptive conditioning to balance motion guidance and semantic refinement across timesteps; and instruction interaction to translate visual drag context into editor-friendly textual guidance. Together, they enable accurate dragging with natural synthesis.

We instantiate these components as follows. \textbf{(i) Region-aware recomposition.} We partition the warped observation into three non-overlapping regions: motion-evidence regions, where warped pixels provide coarse motion cues; disoccluded regions, which should be semantically completed; and preservation regions, where the content should remain unchanged. During sampling, MoRe-Drag maintains separate latent trajectories and recomposes them according to these regional roles. \textbf{(ii) Generation-stage adaptive conditioning.} Since early sampling timesteps determine coarse layout and geometry while later refine appearance and texture details, MoRe-Drag first emphasizes the warped observation to stabilize the drag-induced structure, and then shifts toward the original image and editing instruction for semantic refinement. This design mitigates both warp-induced unnaturalness and instruction-only artifacts. \textbf{(iii) Instruction interaction.} To reduce the interaction burden, we attach a lightweight adapter to the editing model’s MLLM-based text encoder to infer drag-aware instructions from the visual drag annotations. The adapter is activated only for instruction inference and disabled during generation, ensuring that the generative decoder receives embeddings aligned with the original encoder distribution. As illustrated in Fig.~\ref{fig:simple_compare}, our method differs from warp-then-inpaint pipelines such as Inpaint4Drag~\cite{lu2025inpaint4drag} by using the warped image solely as motion evidence rather than as an intermediate edited result, thereby enabling precise drag control with natural synthesis.

\emph{MoRe-Drag} significantly boosts controllable drag editing over text-instruction-only baselines. As shown in Fig.~\ref{fig:teaser}, \emph{MoRe-Drag} produces natural and accurate edits. By grounding the same editor (LongCat-Image-Edit~\cite{DBLP:journals/corr/abs-2512-07584}) with explicit motion evidence, drag accuracy---measured by Mean Distance (MD)---improves from 46.73 to 20.03 on \textsc{DragBench-DR}~\cite{dragdiff} and from 53.62 to 18.54 on \textsc{DragBench-SR}~\cite{sdedrag}.

\section{Related Work}
\textbf{Text-guided image editing.}
Early text-guided image editing methods manipulate pretrained diffusion models through concept learning~\cite{DBLP:conf/iclr/GalAAPBCC23}, attention control~\cite{hertz2022p2p}, and inversion~\cite{mokady2022null}. With advances in flow-based generative modeling~\cite{DBLP:conf/iclr/LipmanCBNL23,DBLP:conf/iclr/LiuG023} and diffusion transformers~\cite{DBLP:conf/iccv/PeeblesX23}, recent editors adopt stronger backbones and multimodal conditioning for open-domain editing~\cite{DBLP:journals/corr/abs-2506-15742,DBLP:conf/aaai/FengMWQCC025}. In particular, MLLM-powered editors, such as FLUX.2~\cite{flux-2-2025}, Qwen-Image-Edit~\cite{wu2025qwenimagetechnicalreport}, LongCat-Image-Edit~\cite{DBLP:journals/corr/abs-2512-07584}, Step1X-Edit~\cite{liu2025step1x-edit}, and FireRed-Image-Edit~\cite{DBLP:journals/corr/abs-2602-13344}, jointly interpret the reference image and editing instruction with multimodal encoders, improving instruction following and appearance preservation. However, text alone remains limited for precise spatial motion control.

\textbf{Drag-based image editing.}
Drag-based editing addresses this limitation through explicit spatial control. DragGAN~\cite{draggan} introduces drag-based editing with point tracking and motion supervision, and DragDiffusion~\citep{dragdiff} extends it to diffusion models. Subsequent methods improve robustness, accuracy, and efficiency~\citep{easydrag,freedrag,noisedrag,gooddrag,stabledrag,DBLP:conf/wacv/ChoiJHH25,DBLP:conf/icml/Xia0SL25,fastdrag,regiondrag,sdedrag}. Inpaint4Drag~\citep{lu2025inpaint4drag} formulates drag editing as pixel-space warp-then-inpaint, whereas MoRe-Drag uses the warped image only as motion evidence to guide a generative editor rather than as an intermediate edit. Recent methods enable drag controllability on MMDiT-based models via explicit correspondence, inversion, region supervision, latent manipulation, or attention alignment~\cite{DBLP:journals/corr/abs-2509-12203,zhou2025dragflow,he2026contextdragprecisedragbasedimage}. In contrast, MoRe-Drag keeps the MMDiT editor unchanged and injects motion externally through conditioning and latent recomposition. Although ContextDrag~\cite{he2026contextdragprecisedragbasedimage} also builds on a pretrained editing model, it relies on attention-level reference injection, whereas \emph{MoRe-Drag} uses motion evidence to better couple precise motion control with natural synthesis.

\section{Methodology}
We first formalize the task of interest. Given an image, user-specified point pairs, and deformation masks, our goal is to produce a natural edit while precisely aligning each handle point with its target point. Recent latent- and pixel-warping methods~\cite{fastdrag,regiondrag,lu2025inpaint4drag} treat the warped outputs as intermediate results, followed by inpainting or restoration to obtain the final drag result (see top of Fig.~\ref{fig:simple_compare}). Text-based methods rely solely on textual instructions to control editing, resulting in limited motion accuracy (see middle of Fig.~\ref{fig:simple_compare}). Moreover, requiring textual instructions further increases the user burden. To address these issues, we introduce \emph{MoRe-Drag}, which combines coarse motion evidence from the warped image with the generative priors of advanced image editors for natural and accurate drag editing (see bottom of Fig.~\ref{fig:simple_compare}).

\begin{figure}[t]
    \centering
    \includegraphics[width=\linewidth]{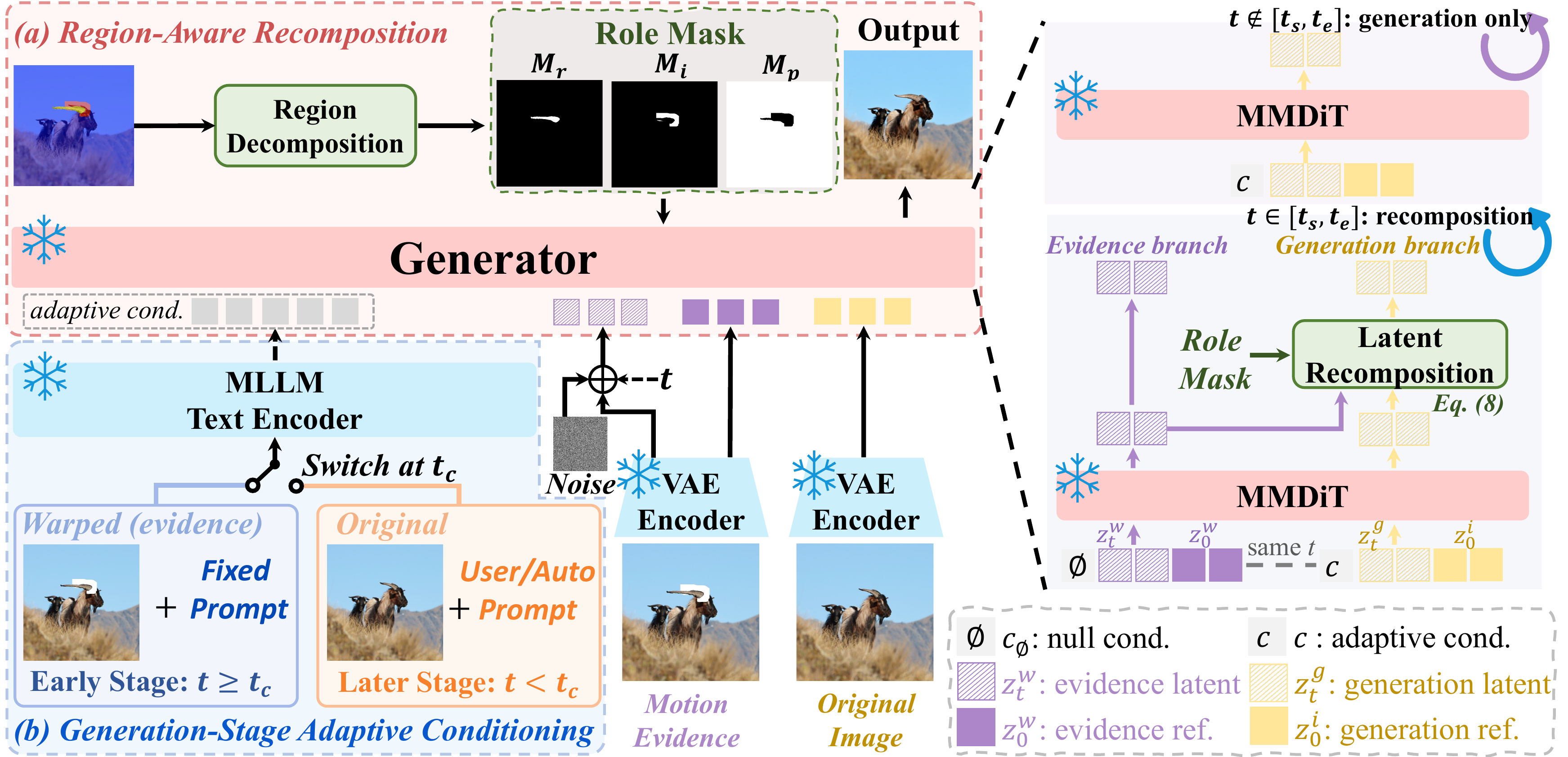}
    
    \caption{\textbf{Overview of \emph{MoRe-Drag}.} The evidence branch and the generation branch are updated by the same MMDiT at the same timestep, and interact only through role-mask recomposition within $[t_s, t_e]$ (right). (a) \textbf{Region-aware recomposition}: the evidence and original image are decomposed into role-specific regions, and the resulting role masks guide latent recomposition (Eq.~\eqref{eq:recon}), so motion cues are injected only into the corresponding regions. (b) \textbf{Generation-stage adaptive conditioning}: at early timesteps $(t \geq t_c)$, the model uses the warped image with a fixed prompt to enforce motion consistency; at later timesteps $(t < t_c)$, it switches to the original image with the user/auto prompt for semantic and appearance refinement. The switch time $t_c$ is set independently of $[t_s, t_e]$.}
    \label{fig:framework}
\end{figure}

\emph{MoRe-Drag} first constructs pixel-space motion evidence. Specifically, following off-the-shelf image warping~\cite{lu2025inpaint4drag}, we interpolate sparse handle displacements 
$\boldsymbol{d}_i=\boldsymbol{t}_i-\boldsymbol{h}_i$ into a dense mask-aware motion field
$\boldsymbol{F}(\boldsymbol{x})=\sum_i\frac{ w_i(\boldsymbol{x})\boldsymbol{d}_i}{\sum_i w_i(\boldsymbol{x})}$,
and bidirectionally warp original image $\boldsymbol{X}_o$ to obtain a coarse evidence image $\boldsymbol{X}_w$, with invalid correspondences marked as disocclusions (detailed in Appendix~\ref{appendix:bidirectional_warping}). Then, as illustrated in Fig.~\ref{fig:framework}, this evidence is incorporated into flow sampling via two components: (i) a region-aware recomposition module that injects motion evidence through region decomposition and latent recomposition (Sec.~\ref{method_1}); (ii) a generation-stage adaptive conditioning strategy that balances motion evidence with semantic refinement across timestep-based condition switch (Sec.~\ref{method_2}). As shown in Fig.~\ref{fig:mllm_finetune}, we fine-tune the MLLM, which also serves as the generator’s text encoder, to support automatic instruction inference (Sec.~\ref{method_3}). For clarity, we provide algorithmic pseudocode in Appendix~\ref{appendix:algorithm}.

\subsection{Region-Aware Recomposition}\label{method_1}
The image warping algorithm~\cite{lu2025inpaint4drag} propagates the sparse handle-to-target displacements across the source region, deforming it according to the indicated motion directions and magnitudes. However, direct warping provides only coarse evidence---it introduces unnatural geometry, local distortions, boundary artifacts, and semantic misalignment, e.g., mouth shapes of the lion and crocodile in Fig.~\ref{fig:teaser}. This raises a key requirement: the editing process should retain the geometric faithfulness of the warp while leveraging the generative prior of the editing model to restore realism and semantic coherence. To this end, we first decompose the warped observation into three functional regions for refinement, hole inpainting, and background preservation (see Fig.~\ref{fig:framework} (a, top)), and then perform latent recomposition to inject the warped image as explicit motion evidence during generation (see Fig.~\ref{fig:framework} (a, right)). This allows the model to follow the user-specified motion while restoring realistic geometry, plausible semantics, and source-region consistency.

\paragraph{Region decomposition.} We obtain the transported target mask $\boldsymbol{M}_t$ by applying the same warp to the source deformation region. As illustrated in the top-left part of Fig.~\ref{fig:framework}, let $\boldsymbol{\Omega}$ be the image domain, direct warping is particularly error-prone near the transported boundary. We therefore augment $\boldsymbol{M}_t$ with a narrow boundary band $\boldsymbol{M}_{\partial t}$ extracted from the contour of the target mask, and define the refinement mask as
\begin{equation}
    \boldsymbol{M}_r = \boldsymbol{M}_t \cup \boldsymbol{M}_{\partial t}.
\end{equation}
To identify the regions that remain uncovered after transport, we take the difference between the source mask and the transported target mask, dilate it to absorb boundary discontinuities, and exclude the refinement mask to avoid overlap:
\begin{equation}
    \boldsymbol{M}_i =
    \operatorname{Dilate}(\boldsymbol{M}_s \setminus \boldsymbol{M}_t)
    \setminus
    \boldsymbol{M}_r,
\end{equation}
where the dilation kernel size is $5$. Finally, all remaining pixels form the preservation mask:
\begin{equation}
\boldsymbol{M}_p =
\boldsymbol{\Omega}
\setminus
(\boldsymbol{M}_r \cup \boldsymbol{M}_i).
\end{equation}
The three masks provide explicit spatial roles for generation. $\boldsymbol{M}_r$ guides refinement of the dragged content, $\boldsymbol{M}_i$ specifies the area to be inpainted, and $\boldsymbol{M}_p$ constrains the non-edited region to remain consistent. 
\paragraph{Latent recomposition.} Given the decomposed masks, we perform latent recomposition by assembling latent features within their corresponding mask regions. As illustrated in the right part of Fig.~\ref{fig:framework}, we maintain two generation trajectories. The yellow trajectory on the right corresponds to conditioned generation branch and is initialized from Gaussian noise: 
\begin{equation}
    \boldsymbol{z}^{g}_{1} \sim \mathcal{N}(\boldsymbol{0}, \boldsymbol{I}).
\end{equation}
We update it by integrating the flow-matching velocity field conditioned on the MLLM text encoder embedding $\boldsymbol{c}$ and reference image $\boldsymbol{z}^{image}_0$, which is obtained by encoding the image with the VAE $\mathcal{E}$:
\begin{equation}
d\boldsymbol{z}^{g}_{t}
=
v_{\theta}(\boldsymbol{z}^{g}_{t}, t, \boldsymbol{c},\boldsymbol{z}^{image}_0)dt,
\quad t:1\rightarrow 0, \quad \text{where} \quad \boldsymbol{z}^{image}_0 = \mathcal{E}(\boldsymbol{X}_o)
\end{equation}
The resulting latent $\boldsymbol{z}^{g}_{t}$ serves as the conditional generation branch for subsequent recomposition.
The purple trajectory on the left corresponds to the evidence branch. We first encode the warped image $\boldsymbol{X}_w$ into the latent space, and then perturb the resulting latent to a noisy state at timestep $t$ for initialization:
\begin{equation}
\boldsymbol{z}^{w}_t=(1-t)\boldsymbol{z}^{w}_0+t\boldsymbol{\epsilon}^{w},
\qquad
\boldsymbol{\epsilon}^{w}\sim\mathcal{N}(\boldsymbol{0},\boldsymbol{I}),\quad
\boldsymbol{z}^{w}_0=\mathcal{E}(\boldsymbol{X}_w),
\end{equation}
The evidence branch follows a null-conditioned reconstruction flow:
\begin{equation}
d\boldsymbol{z}^{w}_{t}
=
v_{\theta}(\boldsymbol{z}^{w}_{t}, t, \boldsymbol{c}_{\emptyset}, \boldsymbol{z}^{w}_0)dt,
\quad t:1\rightarrow 0 .
\end{equation}
The evidence branch provides drag-induced spatial cues as well as reconstruction guidance for the non-edited regions. As shown in the right part of Fig.~\ref{fig:framework}, since both $\boldsymbol{z}^{g}_{t}$ and $\boldsymbol{z}^{w}_{t}$ are defined at the same flow timestep, they can be recomposed in latent space without scale or timestep misalignment.
We resize the refinement, inpainting, and preservation masks to the latent resolution, denoted as $\boldsymbol{m}_r$, $\boldsymbol{m}_i$, and $\boldsymbol{m}_p$. During the timestep interval $\mathcal{T}=[t_s,t_e]$, we recompose the sampling latents:
\begin{equation}
\boldsymbol{z}^{g}_{t'}
\leftarrow
\boldsymbol{m}_i\odot\boldsymbol{z}^{g}_{t'}
+
\boldsymbol{m}_p\odot\boldsymbol{z}^{w}_{t'}
+
\boldsymbol{m}_r\odot
\left(
\lambda(t')\boldsymbol{z}^{g}_{t'}
+
(1-\lambda(t'))\boldsymbol{z}^{w}_{t'}
\right),
\quad t'\in\mathcal{T}.\label{eq:recon}
\end{equation}
Here, $\lambda({t'})$ is a weight controlling the strength of generative refinement. The inpainting region fully relies on the conditioned generation branch to complete the hole, the preservation region adopts the evidence branch to preserve unchanged content, and the refinement region blends both branches to correct geometric artifacts while retaining the drag-induced spatial structure. This region-aware recomposition injects explicit pixel-space motion evidence into the latent sampling trajectory, while leveraging the model's prior to produce natural deformation, plausible completion, and consistent background preservation.
\paragraph{Weight schedule.} We experimented with three timestep-dependent schedules, including constant, linear, and inverse-square decay from $\lambda_{\max}$ to $\lambda_{\min}$. 
Considering its simplicity, efficiency, and stable empirical performance, we use the constant schedule by default. The quantitative comparison of different schedules is reported in Table~\ref{tab:ablation_right}. The formulations of schedules are provided in Appendix~\ref{appendix:weight_schedule}.
\begin{figure}
    \centering
    \includegraphics[width=\linewidth]{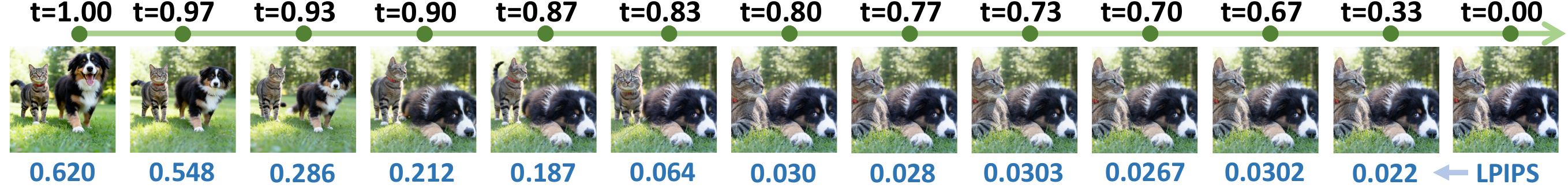}
    \caption{Toy editing from different initial timesteps. Top: starting latent timestep; bottom: LPIPS to the original image.}
    \label{fig:toy_experiment}
\end{figure}
\subsection{Generation-Stage Adaptive Conditioning}\label{method_2}
While region-aware recomposition successfully injects motion evidence into the sampling trajectory, the generator prior still performs semantic refinement based on the source image and editing instruction. As a result, the sampling process may be driven by two competing spatial tendencies: the motion evidence specifies the desired drag target, whereas the generator prior may favor another edit location. When these two tendencies are substantially misaligned, the model attempts to satisfy both, producing duplicated or split edits across different regions, as shown in Fig.~\ref{fig:ablation_1} (only stage 2). This indicates a competition between the editor's semantic prior and the injected motion evidence.

\begin{figure}[t]
    \centering
    \includegraphics[width=\linewidth]{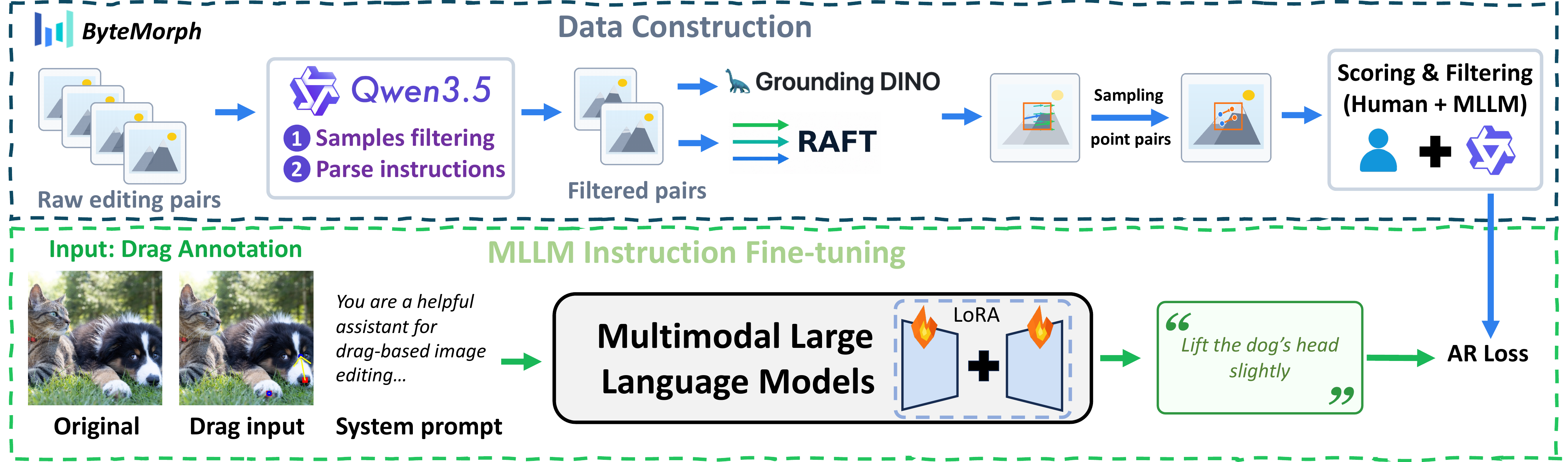}
    \caption{Instruction interaction and data construction. Top: drag annotation construction from \textsc{ByteMorph}. Bottom: LoRA-based MLLM fine-tuning with AR loss, enabling automatic instruction inference from drag annotations.}
    \label{fig:mllm_finetune}
\end{figure}
A common understanding in diffusion/flow-based generation is that early sampling steps tend to establish low-frequency structure and high-level semantics, while later steps increasingly recover high-frequency details~\cite{DBLP:conf/wacv/LeeLG025,DBLP:conf/cvpr/ChoiLSKKY22}. We further conduct a toy editing experiment: given a real image (\(t=0.00\)) and an edit instruction, we perturb the image to different timesteps along the forward flow path and start instruction-conditioned generation from each perturbed state. As shown in the Fig.~\ref{fig:toy_experiment}, even at an early sampling timestep \(t=0.80\), the major image structures and spatial layout have already been determined, while the remaining steps mainly refine appearance details. This observation motivates us to inject motion evidence in the early sampling stage, when spatial decisions are still being formed, and rely on the instruction for later refinement. Thus, as shown in Fig.~\ref{fig:framework}(b), we introduce adaptive conditioning into conditioned generation branch. Specifically, during the early stage, we define the conditioning embedding $\boldsymbol{c}$ as the joint encoding of the warped image (motion evidence) $\boldsymbol{X}_w$, and a fixed prompt $p_{\mathrm{fix}}$:
\begin{equation}
    \boldsymbol{c}^{w}
    =
    \Phi_{\mathrm{mllm}}(\boldsymbol{X}_w, p_{\mathrm{fix}}),
    \qquad
    \boldsymbol{z}^{\mathrm{img},w}_0
    =
    \mathcal{E}(\boldsymbol{X}_w),
\end{equation}
where $\boldsymbol{c}^{w}$ is the early stage condition, and $\Phi_{\mathrm{mllm}}$ denotes MLLM text-encoder. $p_{\mathrm{fix}}$ encourages the model to correct distortions and fill holes in the motion evidence; the full prompt is provided in Box~\ref{box:fixed_prompt}.
Then, switching to the original image and the user/auto prompt during later stage:
\begin{equation}
    \boldsymbol{c}^{o}
    =
    \Phi_{\mathrm{mllm}}(\boldsymbol{X}_o, p),
    \qquad
    \boldsymbol{z}^{\mathrm{img},o}_0
    =
    \mathcal{E}(\boldsymbol{X}_o).
\end{equation}
With this stage-adaptive design, the conditional generation trajectory is sampled as
\begin{equation}
d\boldsymbol{z}^{g}_{t}
=
v_{\theta}
\left(
\boldsymbol{z}^{g}_{t},
t,
\boldsymbol{c}(t),
\boldsymbol{z}^{\mathrm{img}}_0(t)
\right)dt,
\quad \text{where} \left(
\boldsymbol{c}(t),
\boldsymbol{z}^{\mathrm{img}}_0(t)
\right)
=
\begin{cases}
\left(
\boldsymbol{c}^{w},
\boldsymbol{z}^{\mathrm{img},w}_0
\right),
& t \geq t_c,\\[4pt]
\left(
\boldsymbol{c}^{o},
\boldsymbol{z}^{\mathrm{img},o}_0
\right),
& t < t_c .
\end{cases}
\end{equation}
Here, $t_c$ controls the transition from motion-evidence-guided structure formation to source-image-guided semantic refinement. This switching timestep should be properly set to ensure a smooth conditioning transition, balancing motion-evidence injection and semantic refinement. We provide an analysis of its effect in Table~\ref{tab:t_c_analysis}.

This adaptive conditioning strategy complements the region-aware recomposition in Sec.~\ref{method_1} by establishing a dual pathway for motion-evidence injection, while early-stage conditioning guides generation toward the desired motion structure and aligns the model prior with the motion evidence. By first grounding the generation trajectory in the motion evidence and then refining it with the original image and instruction, \emph{MoRe-Drag} improves the alignment between motion evidence and editing instructions while retaining the model’s ability to fix artifacts and synthesize plausible content.

\subsection{Instruction Interaction}\label{method_3}
The above components rely on textual instructions, which places an additional burden on users. To alleviate this, we introduce a no-manual-instruction interface that automatically derives drag-aware instructions from drag-annotation visualizations. Specifically, we reuse the MLLM-based text encoder of the base editor for instruction inference, without introducing any additional standalone model. The inferred instruction is then used for subsequent generation.

Directly using the original MLLM for instruction inference can yield inaccurate intents, such as misidentifying the edited object or introducing semantic changes unsupported by the drag annotations (see Fig.~\ref{fig:instruction}). Therefore, as illustrated in the top part of Fig.~\ref{fig:mllm_finetune}, we construct a high-quality drag-annotation dataset from the released demo split of \textsc{ByteMorph}~\cite{chang2025bytemorph}. Starting from 780K samples, we first remove 560K camera-centric samples using Qwen3.5-9B~\cite{qwen3.5}. We then estimate optical flow with RAFT~\cite{teed2020raft}, localize edited subjects with Grounding DINO~\cite{liu2023grounding}, and cluster regional flow vectors by magnitude and direction to sample handle--target pairs. To further improve annotation quality, we conduct a hybrid human-and-MLLM scoring review of the generated drag annotations, resulting in 68K high-quality samples for subsequent model fine-tuning, detailed in Appendix~\ref{appendix:dataset_construction}.

As shown in the bottom part of Fig.~\ref{fig:mllm_finetune}, we fine-tune the MLLM using LoRA with the standard AR loss, implemented with an open-source Qwen-VL fine-tuning framework\footnote{\url{https://github.com/2U1/Qwen-VL-Series-Finetune}}. The adapter is activated only for instruction inference and disabled during generation, so the MMDiT decoder still receives conditioning features from the original encoder distribution. This bridges geometric interaction and semantic conditioning without altering the generative pathway.

\subsection{Discussion}\label{method_4}
We discuss the motivation behind each design choice. (1) \emph{Why motion evidence?} A pretrained editor offers a strong natural-image prior but no pixel-level motion grounding, whereas the warped image specifies the desired displacement but contains artifacts; we therefore use the warp as motion evidence rather than the final result. (2) \emph{Why an evidence branch?} Relying only on the generation branch may let the prior weaken the displacement, while relying only on the warped image keeps its artifacts. The evidence branch is evolved by the same MMDiT, so its state matches the generation branch in flow timestep, although not necessarily in distribution, and can be fused repeatedly during sampling. (3) \emph{Why region-aware recomposition (RAR)?} Warping leaves transported-but-distorted, disoccluded, and unchanged regions, which require different interactions between the two branches; RAR thus decides \emph{where} the evidence can be trusted, which a single global fusion rule cannot provide. (4) \emph{Why generation-stage adaptive conditioning (GAC)?} RAR decides where the evidence is injected, but the condition decides the direction of each update; GAC thus decides \emph{when} the evidence or the prior should dominate, avoiding the failures of using either condition alone (see Fig.~\ref{fig:ablation_1}). (5) \emph{Why instruction interaction?} The drag geometry is fully specified by the points, masks, and warped image, so the instruction only describes \emph{what} content should be restored, and a short instruction inferred by the reused MLLM suffices (Sec.~\ref{method_3}).

Overall, the evidence branch carries the requested motion, RAR decides \emph{where} this evidence is reliable, GAC decides \emph{when} it should dominate, the generation branch restores natural appearance and missing content, and instruction interaction provides high-level guidance on \emph{what} should be restored.

\begin{table}[t]
\caption{
Results on \textsc{DragBench-SR/DR}. CP and PF are scaled by 100. Metrics of Qwen-Image-Edit and LongCat-Image-Edit are not highlighted, as they are not drag-based methods. ``+ FLUX.1-Fill-dev'' denotes an Inpaint4Drag variant using a stronger inpainting model. (*) indicates results from the original papers, as the models were not publicly available at the time of evaluation. In the Params column, (-) indicates that \emph{MoRe-Drag} keeps the generative backbone frozen.
}
\label{tab:quantitative}
\centering
\renewcommand{\arraystretch}{1}
\footnotesize
\resizebox{\textwidth}{!}{
\begin{tabular}{lc*{4}{c}cc}
\toprule
\multirow{2}{*}{\textbf{Model}}
& \multirow{2}{*}{\textbf{Params}}
& \multicolumn{3}{c}{\textbf{\textsc{DragBench-SR}}}
& \multicolumn{3}{c}{\textbf{\textsc{DragBench-DR}}} \\
\cmidrule(lr){3-5}
\cmidrule(lr){6-8}
&
& \textbf{MD$\downarrow$}
& \textbf{CP$\uparrow$}
& \textbf{PF$\uparrow$}
& \textbf{MD$\downarrow$}
& \textbf{CP$\uparrow$}
& \textbf{PF$\uparrow$} \\
\midrule

\multicolumn{8}{l}{\textit{Test-time Optimization-based Methods}} \\
\midrule
DragDiffusion   & 2.1B   & $41.08_{\pm0.43}$  & $92.75_{\pm0.38}$ & $77.15_{\pm1.49}$ & $31.61_{\pm0.23}$  & $96.46_{\pm0.43}$  & $78.45_{\pm0.95}$ \\
DragNoise       & 2.1B &  $47.64_{\pm1.08}$  &  $79.00_{\pm0.25}$ & $61.56_{\pm0.31}$  & $30.05_{\pm0.01}$  & $91.46_{\pm1.34}$  & $81.02_{\pm0.18}$  \\
GoodDrag        & 2.1B   &  $28.94_{\pm0.40}$  & $94.50_{\pm0.13}$  & $76.50_{\pm0.39}$  & $23.21_{\pm0.12}$  & $95.36_{\pm0.24}$  &  $83.94_{\pm0.04}$ \\
\midrule

\multicolumn{8}{l}{\textit{Optimization-free Methods}}  \\
\midrule
FastDrag             & 2.1B & $23.17_{\pm0.19}$     & $94.00_{\pm0.25}$  & $79.62_{\pm4.38}$ & $29.50_{\pm0.12}$  &  $92.73_{\pm0.73}$ &  $75.60_{\pm1.63}$\\
ContextDrag*         & 12B  & 19.07  &  92.25 & 82.00  & 21.66  & 90.90  & 83.78  \\
Inpaint4Drag         & 2.1B & $19.79_{\pm0.18}$   & $93.00_{\pm0.13}$  &  $81.00_{\pm0.04}$ &  $21.37_{\pm0.32}$  & $88.53_{\pm0.25}$  &  $82.77_{\pm0.76}$\\
\quad + \emph{FLUX.1-Fill-dev} 
                     & 12B   & ${18.96}_{\pm0.18}$   & {$94.69_{\pm0.53}$}  & {$81.41_{\pm1.41}$} & ${21.16}_{\pm0.28 }$  & {$91.70_{\pm0.49}$}  & {$77.19_{\pm1.40}$}  \\
\textcolor{privGray}
{Qwen-Image-Edit}   & \textcolor{privGray}{20B}   & \textcolor{privGray}{ $50.48_{\pm0.12}$}  & \textcolor{privGray}{$94.57_{\pm0.63}$} & \textcolor{privGray}{$83.59_{\pm0.76}$} & \textcolor{privGray}{${47.90}_{\pm0.27}$}  & \textcolor{privGray}{ $92.43_{\pm0.41}$ } & \textcolor{privGray}{$88.17_{\pm0.03}$ } \\
\rowcolor{recaGreen}
\quad + \emph{MoRe-Drag} 
                 & -   & {$23.85_{\pm0.62}$}   & {$90.15_{\pm0.12}$}  & {$70.96_{\pm0.58}$} & {$26.35_{\pm0.17}$}  &  {$87.31_{\pm0.00}$} & {$78.75_{\pm0.89}$}  \\
\textcolor{privGray}
{LongCat-Image-Edit}   & \textcolor{privGray}{6B}   & \textcolor{privGray}{$53.62_{\pm0.09}$ }  & \textcolor{privGray}{ $94.00_{\pm0.50}$ } & \textcolor{privGray}{$92.08_{\pm1.17}$ } & \textcolor{privGray}{ $46.73_{\pm0.24}$ }  & \textcolor{privGray}{ $91.28_{\pm0.55}$ } & \textcolor{privGray}{$91.20_{\pm0.10}$ } \\
\rowcolor{recaGreen}
\quad + \emph{MoRe-Drag} 
                     & -   & $\mathbf{18.54_{\pm0.22}}$  & $\mathbf{96.36_{\pm0.38}}$  & $\mathbf{86.36_{\pm4.06}}$ & $\mathbf{20.03_{\pm0.21}}$ &  $\mathbf{93.15_{\pm0.23}}$ & $\mathbf{84.82_{\pm0.98}}$  \\
\bottomrule
\end{tabular}
}
\end{table}
\begin{figure}
    \centering
    \includegraphics[width=\linewidth]{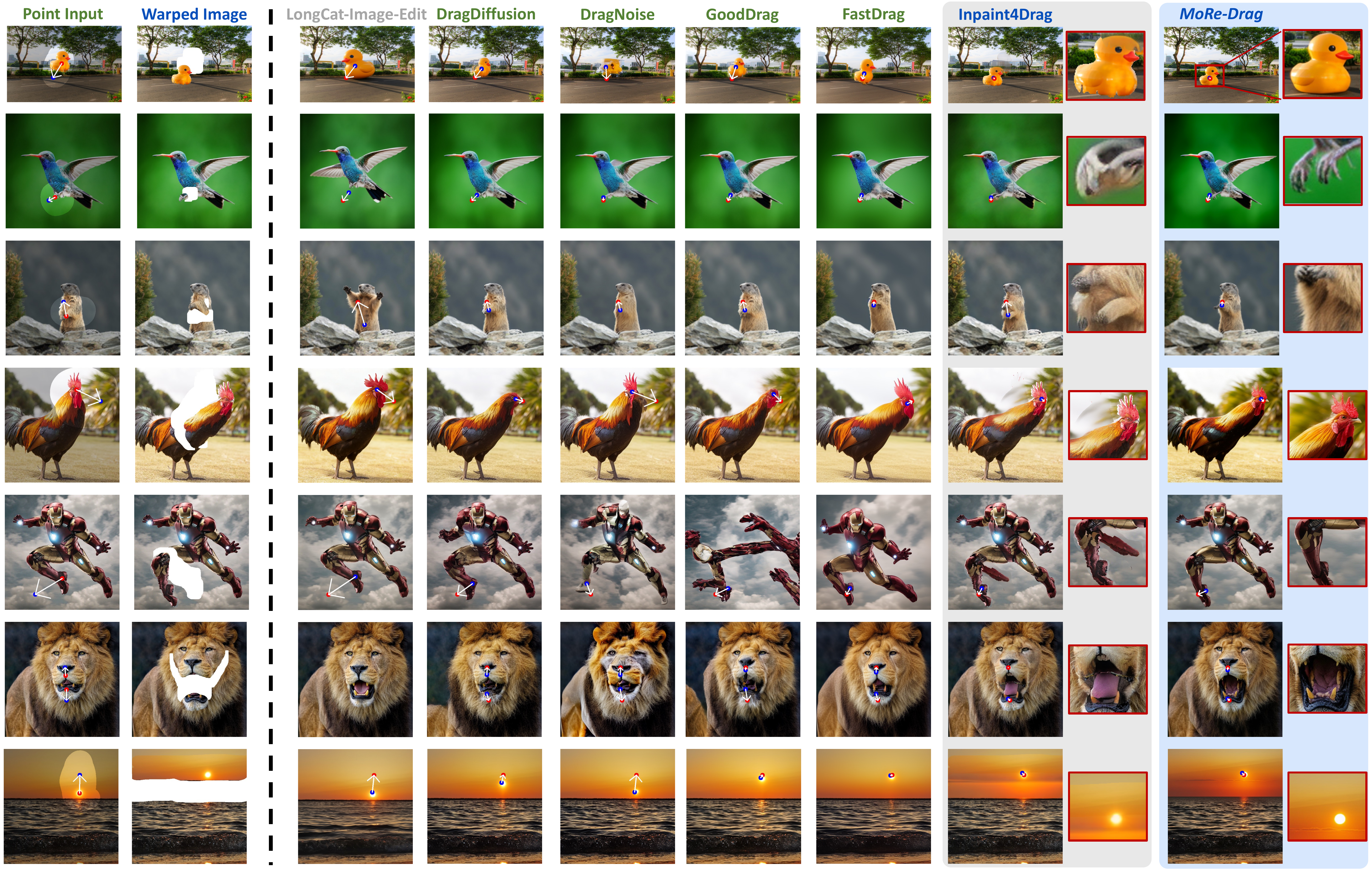}
    \caption{Qualitative comparisons with state-of-the-art drag-based editing models. \textcolor{RoyalBlue}{Blue} and \textcolor{Red}{red} points denote handle and target points, respectively. LongCat-Image-Edit is a text-based editing method; Inpaint4Drag and \emph{MoRe-Drag} take the warped image as input, while the other methods are driven by point inputs. We provide zoomed-in views of the edited regions for the two warped-image-based methods to highlight local structural fidelity. More results are shown in Fig.~\ref{fig:extra_qualitative}}
    \label{fig:quanlitative}
\end{figure}
\section{Experiments}\label{sec:experiment}
\subsection{Implementation Details and Experimental Setup}
\textbf{Implementation Details.} We use LongCat-Image-Edit~\cite{DBLP:journals/corr/abs-2512-07584} as the default base editor for \emph{MoRe-Drag}, and instantiate it on Qwen-Image-Edit~\cite{wu2025qwenimagetechnicalreport} as supplementary. Both editors adopt Qwen2.5-VL-7B~\cite{DBLP:journals/corr/abs-2502-13923} as the text encoder but use different MMDiT backbones. DragDiffusion~\cite{dragdiff}, DragNoise~\cite{noisedrag}, GoodDrag~\cite{gooddrag}, and FastDrag~\cite{fastdrag} take drag points with a mask as input, whereas Inpaint4Drag~\cite{lu2025inpaint4drag} and \emph{MoRe-Drag} use the drag-point-induced warped image. We also include an Inpaint4Drag variant implemented on top of FLUX.1-Fill-dev\footnote{\url{https://huggingface.co/black-forest-labs/FLUX.1-Fill-dev}}. Further details are provided in Appendix~\ref{appendix:implementation_and_training_details}.

\textbf{Experimental Setup.} We evaluate \emph{MoRe-Drag} on \textsc{DragBench-SR}~\cite{sdedrag} and \textsc{DragBench-DR}~\cite{dragdiff}. Since neither benchmark provides natural-language edit instructions, we generate reference instructions with Qwen3.5-9B~\cite{qwen3.5} followed by human correction for evaluation. Unless otherwise specified, \emph{MoRe-Drag} uses inferred instructions produced by our LoRA-adapted MLLM. We report Mean Distance (MD)~\cite{dragdiff}, Concept Preservation (CP)~\cite{DBLP:conf/iclr/PengCTQDBHG0X25}, and Prompt Following (PF)~\cite{DBLP:conf/iclr/PengCTQDBHG0X25}, each averaged over three runs with standard deviations. Details are provided in Appendices~\ref{appendix:benchmark_instruction_generation} and~\ref{appendix:evaluation_metrics}.

\subsection{Comparison with SOTAs} 
\textbf{Quan titative analysis.}
Table~\ref{tab:quantitative} reports the quantitative results. \emph{MoRe-Drag} consistently improves drag precision over the base editors, demonstrating its generality. The performance gap between Qwen-Image-Edit- and LongCat-Image-Edit-based variants mainly stems from the base editors themselves, especially their null-conditioned reconstruction ability; the reconstruction comparison is reported in Appendix~\ref{appendix:base_model}. \emph{MoRe-Drag} based on LongCat-Image-Edit achieves the best drag accuracy on both benchmarks with strong CP and PF scores, showing improved spatial controllability while preserving visual and semantic consistency. Even with the same inputs as ours and a stronger FLUX.1-Fill-dev backbone, Inpaint4Drag still falls short in precise and natural drag editing, highlighting the benefit of using the warped image as motion evidence. Although base editors sometimes obtain slightly higher PF, they are designed for text-guided editing and thus naturally favor this metric; however, they lack precise spatial manipulation ability, leading to much higher MD. Runtime and peak GPU memory comparisons between \emph{MoRe-Drag} and the base editor are reported in Appendix~\ref{appendix:complexity}.

\textbf{Qualitative analysis.} Fig.~\ref{fig:quanlitative} presents qualitative comparisons with representative drag-based editing methods and our base editor (LongCat-Image-Edit). 
\emph{MoRe-Drag} achieves more accurate drag control while producing visually natural edits with better preservation of object texture, shape, and local structures. Compared with prior drag-based methods, \emph{MoRe-Drag} more effectively maintains boundary continuity, preserves fine structures during motion, and avoids distortions, e.g., the first row of Fig.~\ref{fig:quanlitative}. It also enables semantically coherent edits, such as generating realistic mouth details in the lion example and consistently adjusting the sky color and reflection in the sunset case.
\begin{figure}[t]
    \centering

\begin{minipage}[t]{0.49\textwidth}
    \centering
    \captionof{table}{Ablation results on \textsc{DragBench-DR}.}
    \label{tab:ablation_left}
    \begingroup
    \small
    \renewcommand{\arraystretch}{1}
    \setlength{\tabcolsep}{2pt}
    \begin{tabularx}{\linewidth}{>{\raggedright\arraybackslash}Xcccc}
        \toprule
        Variant & MD$\downarrow$  & CP$\uparrow$ & PF$\uparrow$ \\
        \midrule
        \rowcolor{green!5}
        \emph{MoRe-Drag}
        & $\mathbf{20.03_{\pm0.21}}$ &  ${93.15_{\pm0.23}}$ & $\mathbf{84.82_{\pm0.98}}$ \\
        w/o RAR
        & $25.41_{\pm0.23}$ &$92.50_{\pm0.18}$ & $77.14_{\pm1.16}$ \\
        w/o GAC
        & $29.61_{\pm0.33}$ & $\mathbf{93.78_{\pm0.24}}$ & $76.52_{\pm0.67}$ \\
        \bottomrule
    \end{tabularx}
    \endgroup
\end{minipage}
\hfill
\begin{minipage}[t]{0.49\textwidth}
    \centering
    \captionof{table}{Ablation results on \textsc{DragBench-SR}.}
    \label{tab:ablation_right}
    \begingroup
    \small
    \renewcommand{\arraystretch}{1}
    \setlength{\tabcolsep}{2pt}
    \begin{tabularx}{\linewidth}{>{\raggedright\arraybackslash}Xcccc}
        \toprule
        Variant & MD$\downarrow$ & CP$\uparrow$ & PF$\uparrow$ \\
        \midrule
        Constant
        & ${18.54_{\pm0.22}}$  & ${96.36_{\pm0.38}}$  & $\mathbf{86.36_{\pm4.06}}$ \\
        Linear
        & $18.41_{\pm0.03}$ & {$96.25_{\pm0.37}$} &$ 82.25_{\pm0.00}$ \\
        Inv. square
        & $\mathbf{18.23_{\pm0.10}}$ & $\mathbf{96.63_{\pm0.13}}$ & $80.14_{\pm1.86}$ \\
        \bottomrule
    \end{tabularx}
    \endgroup
\end{minipage}

    \vspace{2mm}

    \begin{minipage}[t]{0.56\linewidth}
        \vspace{0pt}
        \centering
        \includegraphics[width=\linewidth]{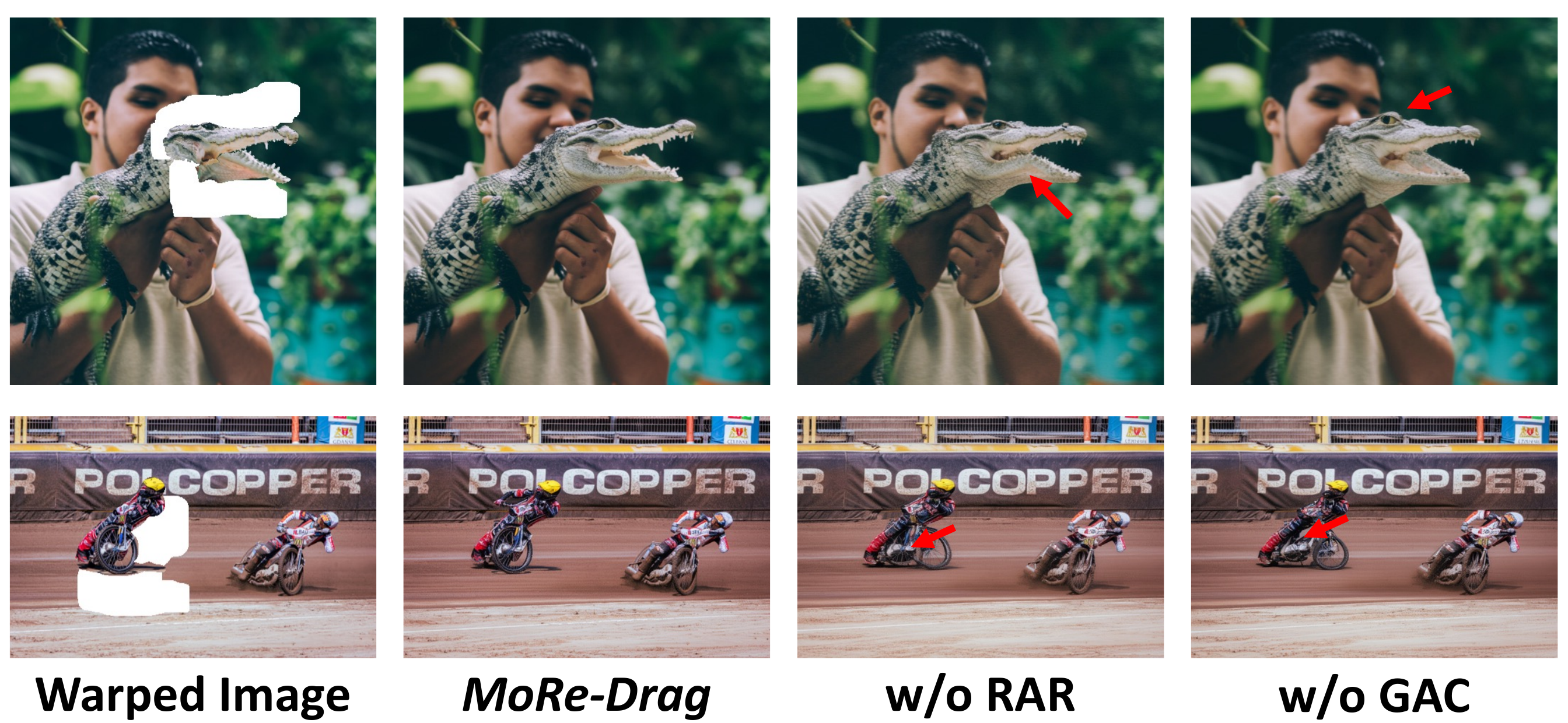}
        \captionof{figure}{Qualitative comparison of \emph{MoRe-Drag} with variants without RAR or GAC.}
        \label{fig:ablation_0}
    \end{minipage}
    \hfill
    \begin{minipage}[t]{0.42\linewidth}
        \vspace{0pt}
        \centering
        \includegraphics[width=\linewidth]{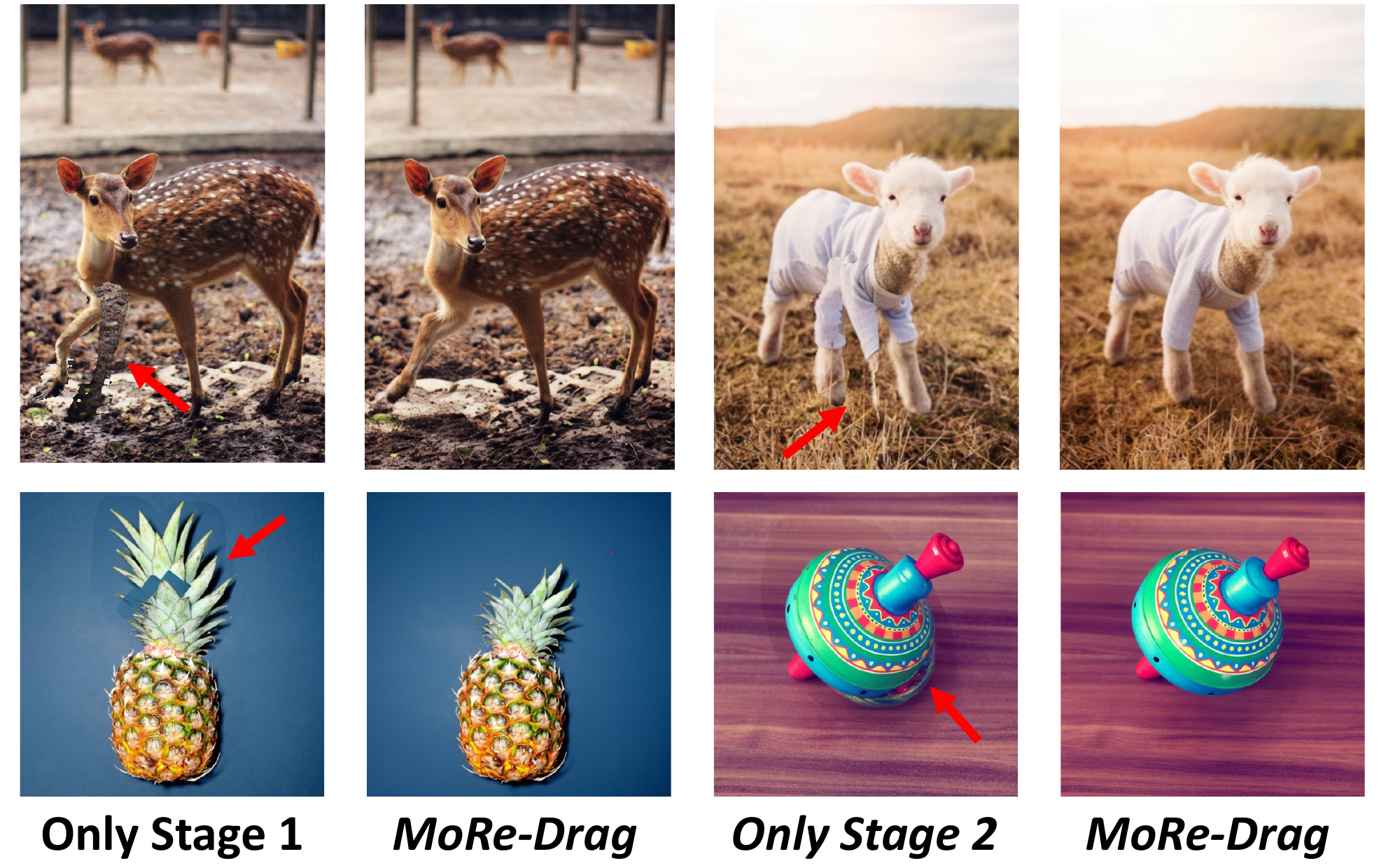}
        \captionof{figure}{
        Effect of generation-stage adaptive conditioning.
        }
        \label{fig:ablation_1}
    \end{minipage}
\end{figure}
\begin{figure}
    \centering
    \includegraphics[width=\linewidth]{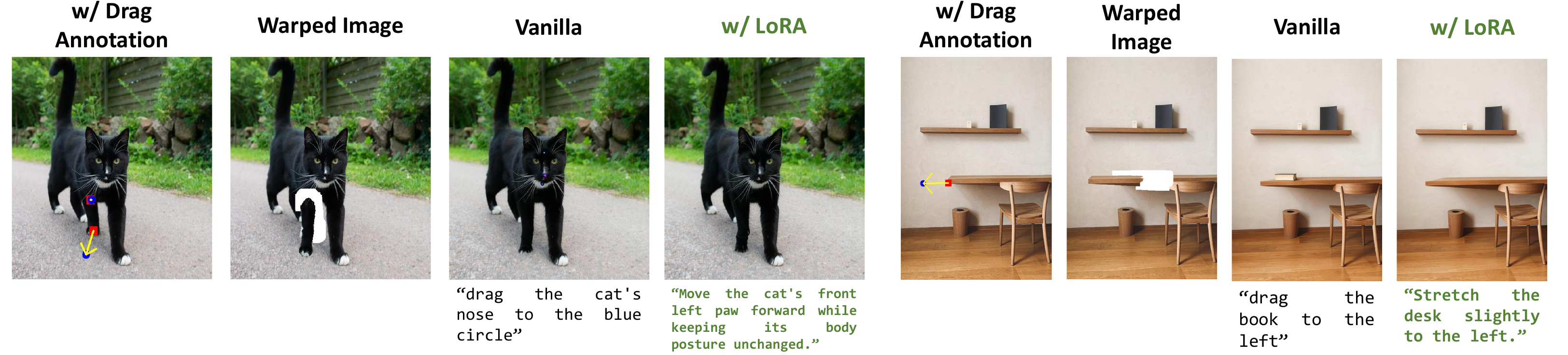}
    \caption{Instruction interaction examples. Compared with vanilla instruction inference, LoRA adaptation produces drag-aware instructions that better match the visual annotations, reducing unintended artifacts and hallucinated content.}
    \label{fig:instruction}
\end{figure}

\textbf{Human preference study.} We conduct a human study with 42 participants, each completing 15 randomized ranking tasks. In each task, the participant is shown the source image with drag annotations and three anonymized results from FastDrag~\cite{fastdrag}, Inpaint4Drag~\cite{lu2025inpaint4drag}, and \emph{MoRe-Drag} in random order, and ranks them by naturalness, drag faithfulness, and preservation, where rank 1 is the best. As shown in Table~\ref{tab:human_study}, \emph{MoRe-Drag} obtains the best average rank on all three criteria. 
\begin{table}[t]
\caption{Human preference study with 42 participants. We report the average rank (lower is better).}
\label{tab:human_study}
\centering
\begin{tabular}{lccc}
\toprule
\textbf{Method} & \textbf{Naturalness$\downarrow$} & \textbf{Drag Faithfulness$\downarrow$} & \textbf{Preservation$\downarrow$} \\
\midrule
FastDrag~\cite{fastdrag}               & 2.52 & 2.30 & 2.48 \\
Inpaint4Drag~\cite{lu2025inpaint4drag} & 2.02 & 2.25 & 1.88 \\
\rowcolor{recaGreen}
\emph{MoRe-Drag}                       & \textbf{1.47} & \textbf{1.45} & \textbf{1.64} \\
\bottomrule
\end{tabular}
\end{table}

\subsection{Ablation Studies}
\textbf{Region-aware recomposition and adaptive conditioning.}
Table~\ref{tab:ablation_left} and Fig.~\ref{fig:ablation_0} show that both Region-Aware Recomposition (RAR) and Generation-Stage Adaptive Conditioning (GAC) are important: removing RAR weakens local deformation control, while removing GAC produces less plausible structures. 
Fig.~\ref{fig:ablation_1} further shows that using only the early-stage condition preserves motion cues but leads to under-refined distortions, whereas using only the later-stage condition can cause competition between semantic editing and injected motion evidence. Since different recomposition weight schedules perform similarly (Table~\ref{tab:ablation_right}), we use the constant schedule by default.

\textbf{Instruction interaction.} Fig.~\ref{fig:instruction} illustrates drag-aware instruction inference. Misaligned instructions can introduce artifacts or unsupported content, whereas LoRA adaptation better captures the drag intent and improves editing faithfulness. Additional comparisons and the inference prompt template are provided in Appendix~\ref{appendix:instruction_interaction_details}.
\section{Conclusion}\label{sec:conclusion}
In this paper, we presented \emph{MoRe-Drag}, a drag-based editing framework that grounds modern editors with motion evidence. Through region-aware recomposition and stage-adaptive conditioning, \emph{MoRe-Drag} improves spatial control while preserving natural and visually coherent image generation results. We also introduced a no-manual-instruction interface for drag-aware intent inference. Overall, our results suggest that pixel-space warping can serve as coarse motion evidence for advanced generative editors, enabling spatially accurate and visually natural drag-based manipulation.

\textbf{Limitations and future work.} Our method is primarily developed for editing backbones that use an MLLM as the text encoder and an MMDiT-style generative decoder. In our experiments, we find that the effectiveness of motion-evidence injection is related to the model's null-conditioned reconstruction ability, where stronger reconstruction generally leads to more reliable editing consistency and concept preservation. Future work may explore improved reconstruction priors and editor-agnostic calibration strategies to make motion grounding more uniformly effective across diverse generative editing backbones.

\begin{ack}
This work was supported by the National Natural Science Foundation of China (Grant Nos. 62676086, 62672109, 62302093, and 52441503), Jiangsu Province Natural Science Fund (BK20230833), the Open Research Fund of the State Key Laboratory of Multimodal Artificial Intelligence Systems under Grant E5SP060116,  the Singapore Ministry of Education (MOE) Academic Research Fund (AcRF) Tier 1 grant (Proposal ID: 23-SIS-SMU-070), and the Lee Kong Chian Fellowship awarded to ZHOU Pan by Singapore Management University. This work was also supported in part by the CIPS-SMP-Zhipu Large Model Fund and was also partially supported by SEU Kunpeng\&Ascend Center of Cultivation. We thank the Big Data Computing Center of Southeast University for providing the facility support on the numerical calculations. Any opinions, findings and conclusions or recommendations expressed in this material are those of the author(s) and do not reflect the views of the Ministry of Education, Singapore. Apart from the funding and support disclosed above, the authors report no financial relationships during the 36 months prior to submission that could be perceived as influencing this work.
\end{ack}

\bibliographystyle{unsrt}
\bibliography{references}

\appendix

\section{Broader Impacts}
\label{appendix:broader_impacts}
\emph{MoRe-Drag} is designed to improve the controllability and usability of interactive image editing. Such capability can benefit creative workflows, visual design, image retouching, education, and AR/VR content creation. Our method reduces the need for manual trial-and-error tuning, enabling more accurate and natural manipulation. However, more controllable and realistic editing tools may also increase the risk of misuse. Precise drag-based manipulation raises potential misuse concerns, including the alteration of visual evidence, the creation of misleading media, and non-consensual edits of personal images. These risks are not unique to our method; however, increased spatial controllability may reduce the effort needed to produce convincing deceptive content. Responsible deployment should therefore incorporate safeguards such as explicit user consent, provenance tracking, watermarking, and authentication of edited images.

\section{Drag-Instruction Dataset Construction}\label{appendix:dataset_construction}
Below, we describe in detail the dataset construction pipeline mentioned in Sec.~\ref{method_3}. \textsc{ByteMorph}\footnote{We use the released demo version, ByteMorph-6M-Demo: \url{https://huggingface.co/datasets/ByteMorph/BM-6M-Demo}.}~\cite{chang2025bytemorph} covers several editing types, including camera zoom, camera motion, object motion, human motion, and human-object interaction. Since our method focuses on drag-based editing, we design an automated data processing pipeline to extract samples with reliable local motion, mine dense drag correspondences, filter noisy annotations, and construct spatial candidate regions. The overall data construction pipeline is summarized in Table~\ref{tab:data_pipeline}. Starting from \textsc{ByteMorph}, the motion filtering stage retains approximately 220K samples, and the final score-based filtering produces approximately 68K high-quality drag-instruction samples.

\paragraph{(i) Motion-edit filtering.}
We first filter \textsc{ByteMorph} to keep only samples involving local motion edits. Specifically, we use Qwen3.5-9B~\cite{qwen3.5} as a structured instruction parser. Given an editing instruction, the model determines whether the instruction is motion-related, whether it involves global camera changes, the motion type, and the entities being edited. We retain a sample only if it describes human or object motion, and discard samples involving global camera changes. The extracted edited entities are further used in the subsequent step to obtain masks for the editing subjects. This stage retains approximately 220K motion-related samples from the original dataset.

\paragraph{(ii) Drag annotation mining.}
For each retained sample, we automatically extract drag annotations from the source--target image pair. We first localize the edited entities obtained from the instruction parser. Grounding DINO~\cite{liu2023grounding} is used to detect entity boxes in the source image, and SAM~\cite{DBLP:conf/iccv/KirillovMRMRGXW23} converts the detected boxes into pixel-level masks. We then compute bidirectional optical flow between the source and target images using RAFT~\cite{teed2020raft}. A pixel is considered a valid motion candidate if the magnitudes of both its forward flow and the corresponding backward flow exceed $1.5$ pixels.

The candidate motion mask is defined as the intersection of the entity mask and the valid motion mask. We remove a $4$-pixel image border and discard connected components with fewer than $64$ pixels. Large components are further partitioned by applying $k$-means clustering to their flow vectors, which separates regions with different motion directions. We retain at most $8$ motion regions per sample. Within each region, handle points are sampled on a stratified grid, at a density of approximately one point per $16\times16$ pixels. The number of handle points is capped at $8$ per region and $64$ per sample. For each handle point $(x_h,y_h)$, the corresponding target point is obtained by adding the forward optical-flow vector:
\[
(x_t,y_t) = (x_h + \Delta x, y_h + \Delta y).
\]
We discard target points outside the image boundary and remove samples that contain no valid drag points.

\paragraph{(iii) Drag quality scoring and filtering.}
The automatically mined drag annotations may still contain noisy correspondences or motion cues that are not semantically aligned with the intended edit. We therefore score the annotations with a VLM, using human inspection to calibrate the scoring prompt and spot-check the retained samples. For each sample, Qwen3.5-9B~\cite{qwen3.5} is given three images: the source image, the source image overlaid with drag guidance, and the target image. The drag visualization marks handle points in red, target points in blue, and drag directions with arrows.

The VLM scores each sample along three dimensions: camera motion, instruction alignment, and target alignment. The camera-motion score penalizes global camera changes in the source--target pair; the instruction-alignment score measures whether the drag guidance matches the editing instruction; and the target-alignment score measures whether the drag guidance explains the visible transformation from the source image to the target image. Each score is an integer in $\{0,1,2,3\}$, with higher values indicating better quality. The final score is the normalized average:
\[
S_{\mathrm{avg}}
=
\frac{1}{3}
\left(
\frac{S_{\mathrm{cam}}}{3}
+
\frac{S_{\mathrm{ins}}}{3}
+
\frac{S_{\mathrm{tgt}}}{3}
\right).
\]
We discard samples with $S_{\mathrm{avg}} < 0.6$ and remove obvious failure cases through human spot-checking. This score-based filtering step yields approximately 68K high-quality drag-annotated samples. The prompt used for VLM-based drag quality scoring is shown in Box~\ref{box:drag_quality_prompt}.

\begin{table}[t]
\centering
\caption{Summary of the drag-instruction dataset construction pipeline.}
\label{tab:data_pipeline}
\resizebox{\linewidth}{!}{
\begin{tabular}{l l l c}
\toprule
Stage & Main models & Key settings & Size \\
\midrule
Raw dataset
& -
& -
& 780K \\
Motion-edit filtering 
& Qwen3.5-9B 
& local human/object motion; no global camera change 
& 220K \\
Drag annotation mining 
& Grounding DINO, SAM, RAFT 
& flow threshold $1.5$
& 220K \\
Quality scoring and filtering
& Qwen3.5-9B + Human
& alignment scores in $\{0,1,2,3\}$; keep $S_{\mathrm{avg}} \geq 0.6$
& 68K \\
\bottomrule
\end{tabular}
}
\end{table}
\section{Implementation, Evaluation, and Training Details}
\label{appendix:implementation_and_training_details}
\subsection{Implementation Details}\label{appendix:implementation}
We implement \emph{MoRe-Drag} on LongCat-Image-Edit~\cite{DBLP:journals/corr/abs-2512-07584} and Qwen-Image-Edit~\cite{wu2025qwenimagetechnicalreport}. Both models adopt Qwen2.5-VL-7B~\cite{DBLP:journals/corr/abs-2502-13923} as the MLLM-based text encoder, while using MMDiT generators with 6B and 20B parameters, respectively. All evaluation experiments are conducted on NVIDIA RTX 4090 GPUs. Unless otherwise specified, we use 30 sampling steps. For \textsc{DragBench-SR}~\cite{sdedrag}, we use a CFG scale of 3, apply region-aware latent recomposition from step 1 to step 16, and switch generation-stage adaptive conditioning at step 12. For \textsc{DragBench-DR}, we use a CFG scale of 2, apply region-aware latent recomposition from step 1 to step 14, and switch generation-stage adaptive conditioning at step 11. For the recomposition weight, we set $\lambda_{\max}=0.5$ and $\lambda_{\min}=0.3$.
\subsection{Training Details}\label{appendix:training}
For drag-aware instruction inference, we fine-tune the Qwen2.5-VL-7B text encoder on our filtered 68K drag-instruction dataset. The fine-tuning is performed with LoRA on two NVIDIA RTX PRO 6000 GPUs for 15 GPU-hours. We freeze the vision tower, LLM backbone, and multimodal merger, and train LoRA adapters with rank 32, alpha 64, and dropout 0.05. The model is trained for 3 epochs using a global batch size of 80, with a per-device batch size of 40 and no gradient accumulation. We use bfloat16 precision and DeepSpeed ZeRO-2 setting. The learning rate is set to $2\times10^{-4}$ with weight decay 0.1, a warmup ratio of 0.03, and a cosine learning-rate schedule. The input image resolution is controlled by setting the minimum and maximum visual tokens to $256\times 28\times 28$ and $1280\times 28\times 28$, respectively.
\subsection{Formulation of Weight Schedule}\label{appendix:weight_schedule}
We provide the formulations of the weight schedules used in latent recomposition (Sec.~\ref{method_1}). Unless otherwise specified, we adopt the constant schedule as the default setting.
\begin{equation}
\begin{aligned}
\textit{Constant}:~\lambda(t_i)&=\lambda_{\max},\quad
\textit{Linear}:~\lambda(t_i)=\lambda_{\max}+(\lambda_{\min}-\lambda_{\max})\rho_i,\\
\textit{Inverse square}:~\lambda(t_i)&=\lambda_{\max}+(\lambda_{\min}-\lambda_{\max})\rho_i^2.
\end{aligned}
\end{equation}
\subsection{Details of Bidirectional Warping}\label{appendix:bidirectional_warping}
We adopt the bidirectional pixel-space warping strategy from Inpaint4Drag~\cite{lu2025inpaint4drag} to construct the warped observation used as motion evidence. Given a source image $\boldsymbol{X}$, a manipulation mask $\boldsymbol{M}_{s}$, and handle--target pairs $\{(\boldsymbol{h}_i,\boldsymbol{t}_i)\}_{i=1}^{K}$, the goal is to propagate sparse user-specified displacements to dense pixel motions within the editable region. For each handle point, the displacement vector is defined as
\begin{equation}
    \boldsymbol{d}_i = \boldsymbol{t}_i - \boldsymbol{h}_i .
\end{equation}
A dense motion field $\boldsymbol{F}(\boldsymbol{x})$ is then estimated by interpolating the sparse displacements over pixels $\boldsymbol{x}\in\boldsymbol{M}_{s}$, with nearby handles contributing more strongly:
\begin{equation}
    \boldsymbol{F}(\boldsymbol{x})
    =
    \sum_{i=1}^{K}\frac{ w_i(\boldsymbol{x}) \boldsymbol{d}_i}
    {\sum_{i=1}^{K} w_i(\boldsymbol{x})},
    \qquad
    w_i(\boldsymbol{x}) =
    \frac{1}{\|\boldsymbol{x}-\boldsymbol{h}_i\|+\epsilon}.
\end{equation}
Pixels outside the manipulation mask are kept unchanged.

Following Inpaint4Drag~\cite{lu2025inpaint4drag}, we apply bidirectional warping to better preserve transported content while identifying disoccluded regions. Forward warping maps each source pixel to its displaced location:
\begin{equation}
    \boldsymbol{x}' = \boldsymbol{x} + \boldsymbol{F}(\boldsymbol{x}),
    \qquad \boldsymbol{x}\in\boldsymbol{M}_{s}.
\end{equation}
Since forward warping may create holes or overlaps, we further construct a backward mapping for pixels in the transformed target region. For each target pixel $\boldsymbol{x}'$, we find its $N$ nearest pixels $\{\boldsymbol{x}'_{i}\}_{i=1}^{N}$ that have valid correspondences from the forward warping step, where $\boldsymbol{x}_{i}$ denotes the corresponding source position of $\boldsymbol{x}'_{i}$. The source position of $\boldsymbol{x}'$ is then estimated by locally interpolating the inverse displacement:
\begin{equation}
    \boldsymbol{x}_{s}(\boldsymbol{x}')
    =
    \boldsymbol{x}'
    +
    \sum_{i=1}^{N}
    w_i(\boldsymbol{x}')
    \left(
        \boldsymbol{x}_{i} - \boldsymbol{x}'_{i}
    \right),
\end{equation}
where $w_i(\boldsymbol{x}')$ are normalized inverse-distance weights computed from the distance between $\boldsymbol{x}'$ and $\boldsymbol{x}'_{i}$. We only keep valid mapping pairs whose source and target coordinates lie inside the image domain,
\begin{equation}
    \boldsymbol{x}_{s}(\boldsymbol{x}'), \boldsymbol{x}'
    \in [0,W) \times [0,H),
\end{equation}
where $W$ and $H$ denote the image width and height. The warped image is then obtained by 
\begin{equation}
    \boldsymbol{X}_{w}(\boldsymbol{x}')
    =
    \boldsymbol{X}\!\left(\boldsymbol{x}_{s}(\boldsymbol{x}')\right).
\end{equation}
The resulting warped image $\boldsymbol{X}_{w}$ provides a coarse observation of the desired drag motion.

\begin{figure}[t]
    \centering
    \includegraphics[width=\linewidth]{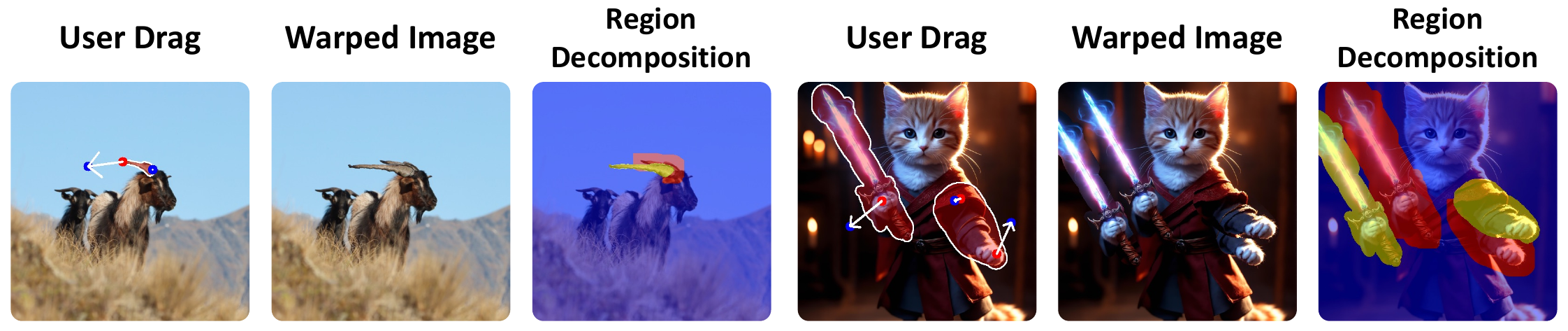}
    \caption{
    Illustration of bidirectional warping and mask decomposition.
    Given handle--target drag annotations, bidirectional warping produces a warped observation that provides pixel-space motion evidence.
    We decompose the result into preservation, refinement, and inpainting regions for region-aware latent recomposition:
    \textcolor{blue}{blue} denotes the anchor preservation $\boldsymbol{M}_p$,
    \textcolor{yellow!70!black}{yellow} denotes the refinement mask $\boldsymbol{M}_r$,
    and \textcolor{red}{red} denotes the inpainting mask $\boldsymbol{M}_i$.
    }
    \label{fig:bidirectional_warping_demo}
\end{figure}

Fig.~\ref{fig:bidirectional_warping_demo} visualizes the intermediate representation derived from drag annotations. The user-specified drag input is converted into a warped observation that serves as pixel-space motion evidence. Based on this observation, we construct three functional masks for region-aware latent recomposition: a preservation mask for unchanged content, a refinement mask for transported content, and an inpainting mask for invalid regions.

This warping result is not used as the intermediate edited image. Instead, \emph{MoRe-Drag} treats $\boldsymbol{X}_{w}$ as coarse motion evidence that provides drag-relevant cues about where and how the content should move. The subsequent region-aware recomposition and adaptive conditioning stages repair artifacts introduced by warping and synthesize visually plausible content. For additional implementation details of the bidirectional warping procedure, we refer readers to Inpaint4Drag~\cite{lu2025inpaint4drag}.
\subsection{Benchmark Instruction Generation}\label{appendix:benchmark_instruction_generation}
\textsc{DragBench-SR}~\cite{sdedrag} and \textsc{DragBench-DR}~\cite{dragdiff} provide source images, manipulation masks, and handle--target point pairs, but do not include the natural-language editing instructions needed for our instruction-based evaluation metrics. We therefore generate one edit prompt for each sample with a vision-language model. For each sample, we pair the original source image with an annotated version in which handle points are marked as red squares, target points as blue circles, and handle-to-target displacements as yellow arrows. We feed this two-image input to Qwen3.5-9B~\cite{qwen3.5}, together with a structured intent-parsing prompt (Box~\ref{box:benchmark_prompt}), to obtain a concise instruction that captures the semantic intent of the drag operation. The generated instructions are manually checked and corrected when necessary. The final instruction is written back to the sample metadata for instruction-based evaluation, while the original drag points and masks are kept for drag-based editing and metric computation.
\subsection{Evaluation Metrics}\label{appendix:evaluation_metrics}
Following DragDiffusion~\cite{dragdiff}, we use Mean Distance (MD) to evaluate drag precision. MD measures the average Euclidean distance between the tracked handle points in the edited image and the user-specified target points; lower values indicate more accurate drag control. Specifically, for each handle--target pair, we track the handle location after editing, compute its distance to the target point, and average the distances over all point pairs and samples.

We do not report Image Fidelity (IF), as prior work has noted that it is not well suited for evaluating identity preservation in drag-based editing~\cite{lu2025inpaint4drag,DBLP:journals/corr/abs-2509-12203}. For drag-based editing, valid geometric transformations may inevitably increase LPIPS despite preserving the intended identity, making IF an unreliable proxy for fidelity in this setting. We therefore adopt the vision-language-model-based metrics from DreamBench++~\cite{DBLP:conf/iclr/PengCTQDBHG0X25}, namely Concept Preservation (CP) and Prompt Following (PF), which are designed to better capture semantic consistency and instruction alignment.

Following DreamBench++, we use GPT-4o as the evaluator for both CP and PF. CP is computed under the original DreamBench++ setting to assess whether the edited image preserves the main visual concept and identity of the source content. For PF, since our task is image editing rather than text-to-image generation, we adapt the original prompt-following protocol by providing the evaluator with a triplet consisting of the source image, the edited image, and the corresponding edit instruction. The evaluator is asked to determine whether the edited image follows the instruction while preserving visual content that should remain unchanged. The PF score ranges from 0 to 4, with higher values indicating better instruction alignment and content preservation. The prompt used for PF evaluation is shown in Box~\ref{box:pf_prompt}.

\begin{algorithm}[t]
\caption{\emph{MoRe-Drag}}
\label{alg:more_drag}
\begin{algorithmic}[1]
\Require Source image $\boldsymbol{X}_o$, warped image $\boldsymbol{X}_w$, drag masks $(\boldsymbol{m}_r,\boldsymbol{m}_i,\boldsymbol{m}_p)$, instruction $p$, fixed restoration prompt $p_{\mathrm{fix}}$, recomposition interval $\mathcal{T}$, switching timestep $t_c$.
\State Encode source and warped images: $\boldsymbol{z}^{o}_{0}=\mathcal{E}(\boldsymbol{X}_o)$, $\boldsymbol{z}^{w}_{0}=E(\boldsymbol{X}_w)$.
\State Initialize generation latent $\boldsymbol{z}^g_1\sim\mathcal{N}(0,\boldsymbol{I})$ and evidence latent $\boldsymbol{z}^w_t$ from the flow perturbation of $\boldsymbol{z}^w_0$.
\For{each timestep $t:1\rightarrow 0$}
    \If{$t \geq t_c$}
        \State Use warped-image condition $(\boldsymbol{c}(t),\boldsymbol{z}^{\mathrm{img}}_0(t))=(\Phi_{\mathrm{mllm}}(\boldsymbol{X}_w,p_{\mathrm{fix}}), \mathcal{E}(\boldsymbol{X}_w))$.
    \Else
        \State Use original-image condition $(\boldsymbol{c}(t),\boldsymbol{z}^{\mathrm{img}}_0(t)) =(\Phi_{\mathrm{mllm}}(\boldsymbol{X}_o,p),\mathcal{E}(\boldsymbol{X}_o))$.
    \EndIf
    \State Update the generation trajectory with one MMDiT forward pass:
    \Statex \hspace{1.5em} $\boldsymbol{z}^g_{t-\Delta t}\leftarrow \mathrm{Step}(\boldsymbol{z}^g_t, \boldsymbol{v}_\theta(\boldsymbol{z}^g_t,t,\boldsymbol{c}(t),\boldsymbol{z}^{\mathrm{img}}_0(t)))$.
    \If{$t\in\mathcal{T}$}
        \State Update the evidence trajectory with a null-condition MMDiT forward pass:
        \Statex \hspace{1.5em} $\boldsymbol{z}^w_{t-\Delta t}\leftarrow \mathrm{Step}(\boldsymbol{z}^w_t, \boldsymbol{v}_\theta(\boldsymbol{z}^w_t,t,\boldsymbol{c}_\emptyset,\boldsymbol{z}^w_0))$.
        \State Recompose the generation latent:
        \Statex \hspace{1.5em} $\boldsymbol{z}^g_{t-\Delta t}\leftarrow
        \boldsymbol{m}_i\odot \boldsymbol{z}^g_{t-\Delta t}+\boldsymbol{m}_p\odot \boldsymbol{z}^w_{t-\Delta t}
        +\boldsymbol{m}_r\odot(\lambda \boldsymbol{z}^g_{t-\Delta t}+(1-\lambda)\boldsymbol{z}^w_{t-\Delta t})$.
    \EndIf
\EndFor
\State Decode $\boldsymbol{z}^g_0$ to obtain the edited image.
\end{algorithmic}
\end{algorithm} 
\section{Algorithmic Details of MoRe-Drag}
\label{appendix:algorithm}
Algorithm~\ref{alg:more_drag} summarizes the inference procedure of \emph{MoRe-Drag}. The input instruction $p$ can be either provided by the user or automatically inferred from drag annotations using the fine-tuned MLLM text encoder of the editor (Sec.~\ref{method_3}). The method maintains two latent trajectories during sampling: an instruction-conditioned generation trajectory and a null-condition evidence reconstruction trajectory. The generation trajectory is updated at every sampling step using the adaptive image-text condition (Sec.~\ref{method_2}). The evidence trajectory is updated only within the recomposition interval $\mathcal{T}=[t_s,t_e]$, where the warped observation is injected as motion evidence (Sec.~\ref{method_1}).

\section{Extra Results}\label{appendix:extra_results}
\subsection{Quantitative Results}\label{appendix:extraquantitative_results}
\begin{table}[h]
\caption{Comparison with RegionDrag and DragFlow on \textsc{DragBench-SR/DR}. CP and PF are scaled by 100. \emph{MoRe-Drag} uses LongCat-Image-Edit as the base editor.}
\label{tab:extra_baselines}
\centering
\footnotesize
\begin{tabular}{lcccccc}
\toprule
\multirow{2}{*}{\textbf{Model}}
& \multicolumn{3}{c}{\textbf{\textsc{DragBench-SR}}}
& \multicolumn{3}{c}{\textbf{\textsc{DragBench-DR}}} \\
\cmidrule(lr){2-4}
\cmidrule(lr){5-7}
& \textbf{MD$\downarrow$} & \textbf{CP$\uparrow$} & \textbf{PF$\uparrow$}
& \textbf{MD$\downarrow$} & \textbf{CP$\uparrow$} & \textbf{PF$\uparrow$} \\
\midrule
RegionDrag~\cite{regiondrag}      & 22.73 & 77.25 & 65.75 & 26.36 & 76.10 & 63.05 \\
DragFlow~\cite{zhou2025dragflow}  & 36.87 & 76.75 & 60.00 & 40.59$^\dagger$ & 75.49 & 65.98 \\
\rowcolor{recaGreen}
\emph{MoRe-Drag}                  & \textbf{18.54} & \textbf{96.36} & \textbf{86.36} & \textbf{20.03} & \textbf{93.15} & \textbf{84.82} \\
\bottomrule
\end{tabular}
\par\vspace{2pt}
{\scriptsize\itshape $^\dagger$The original DragFlow paper reports an MD of 31.59 on \textsc{DragBench-DR}.\par}
\end{table}

We further compare \emph{MoRe-Drag} with RegionDrag~\cite{regiondrag} and DragFlow~\cite{zhou2025dragflow} using their released implementations, where the target-region centroids required by DragFlow are derived from the benchmark annotations. As shown in Table~\ref{tab:extra_baselines}, \emph{MoRe-Drag} outperforms both methods on all metrics, and its MD remains lower than that reported in the DragFlow paper.

\subsection{Qualitative Results}\label{appendix:extra_qualitative_results}
\begin{figure}[h]
    \centering
    \includegraphics[width=\linewidth]{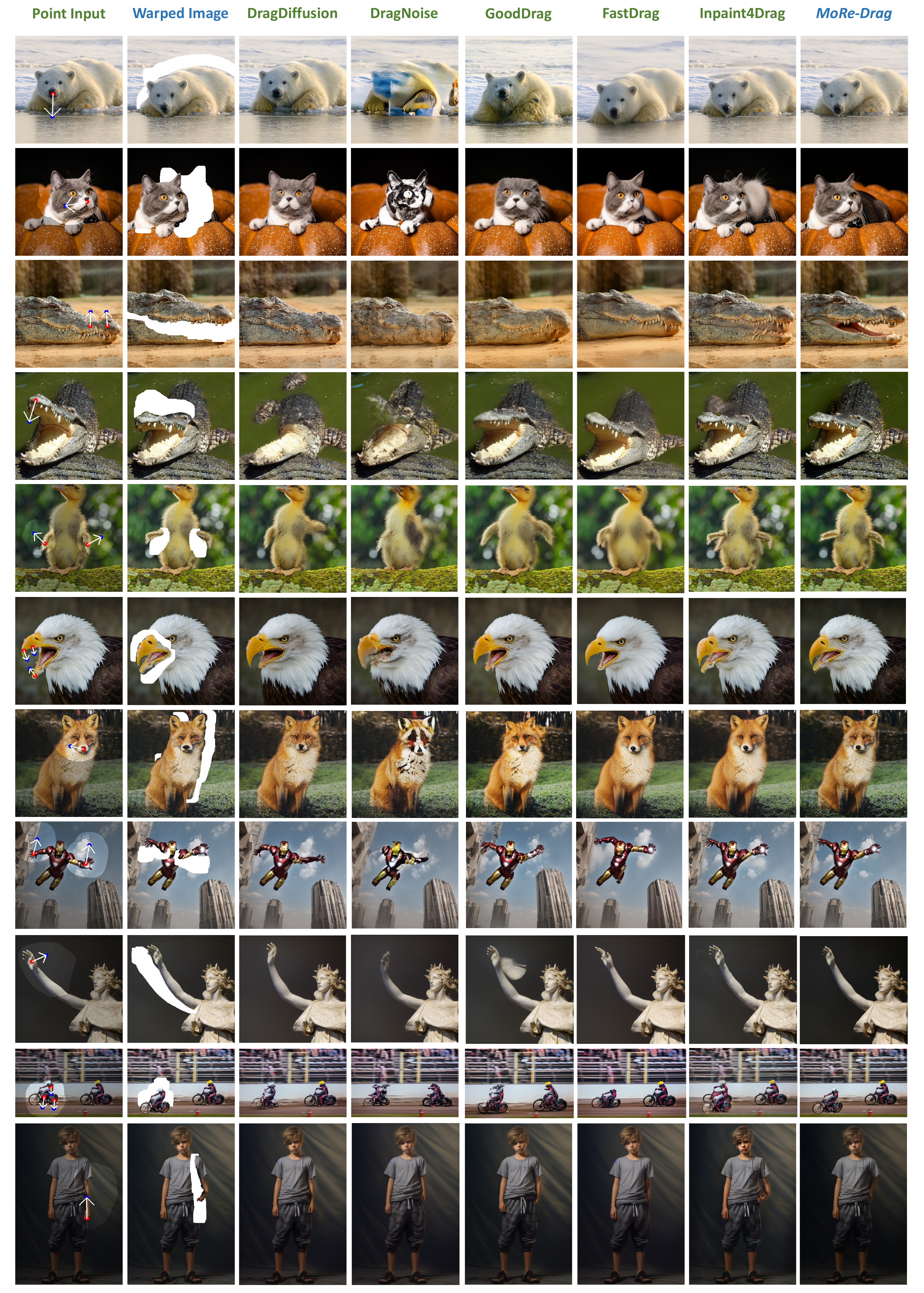}
\caption{Additional qualitative comparisons on diverse drag-based editing examples. \emph{MoRe-Drag} faithfully follows the specified drag motion while preserving object structure, texture details, and background consistency.}
\label{fig:extra_qualitative}
\end{figure}

Fig.~\ref{fig:extra_qualitative} presents additional qualitative comparisons across diverse drag-based editing scenarios, including object translation, pose adjustment, and human/object motion. Existing drag-based methods often suffer from a precision--naturalness trade-off. Optimization-based methods, such as DragDiffusion~\cite{dragdiff}, DragNoise~\cite{noisedrag}, and GoodDrag~\cite{gooddrag}, lack a sufficiently strong mechanism for motion injection, which limits their ability to perform high-precision edits. Optimization-free methods also exhibit clear limitations: FastDrag~\cite{fastdrag} performs warping in the latent space followed by generative refinement, which can lead to detail loss and inaccurate edits, as shown in Rows 3 and 5 of Fig.~\ref{fig:extra_qualitative}. Inpaint4Drag~\cite{lu2025inpaint4drag} uses the warped image as an intermediate result and can achieve relatively accurate motion in some cases, but it cannot avoid distortions and unnatural artifacts introduced by warping, as observed in Rows 1, 3, 5, and 8 of Fig.~\ref{fig:extra_qualitative}. In contrast, \emph{MoRe-Drag} uses the warped image as motion evidence to provide deformation constraints for the editing model, enabling edits that are both accurate and natural. For example, in Row 3 of Fig.~\ref{fig:extra_qualitative}, where the crocodile's mouth is opened, \emph{MoRe-Drag} satisfies the drag constraints while generating realistic inner-mouth content. In Row 8, \emph{MoRe-Drag} successfully adjusts Iron Man's pose while preserving a more natural and plausible arm structure, whereas Inpaint4Drag produces visible arm distortions. These examples further demonstrate the effectiveness of our method.
\begin{table}[t]
\centering
\caption{Hyperparameter analysis on \textsc{DragBench-SR}.}
\label{tab:hyperparameter_analysis}
\resizebox{0.75\linewidth}{!}{
\begin{tabular}{cccc}
\toprule
\multirow{2}{*}{Setting (RAR steps, GAC switch)}
& \multicolumn{3}{c}{\textsc{DragBench-SR}} \\
\cmidrule(lr){2-4}
& MD$\downarrow$ & CP$\uparrow$ & PF$\uparrow$ \\
\midrule
(6, 2)   & $37.00_{\pm0.48}$ & $92.75_{\pm0.00}$ & $69.72_{\pm0.28}$ \\
(11, 9)  & $21.09_{\pm0.20}$ & $95.75_{\pm0.25}$ & $80.63_{\pm0.38}$ \\
(16, 12) & ${18.54_{\pm0.22}}$  & ${96.36_{\pm0.38}}$  & ${86.36_{\pm4.06}}$ \\
(21, 14) & $18.43_{\pm0.31}$ & $95.87_{\pm0.88}$ & $83.50_{\pm2.36}$ \\
(30, 20) &$17.98_{\pm0.37}$ & $94.25_{\pm1.00}$ & $80.75_{\pm1.61}$ \\
\bottomrule
\end{tabular}
}
\end{table}

\subsection{Comparison with Design Alternatives}\label{appendix:design_alternatives}
Table~\ref{tab:design_alternatives} compares \emph{MoRe-Drag} with simpler alternatives. To check whether RAR is needed, we replace it with global blending, which keeps the same two branches and blending weight but mixes the whole latent without role masks. This variant is even worse than removing RAR, showing that the benefit of RAR comes from handling each region separately.

We then examine the evidence branch. One alternative re-noises the warped latent to the current timestep, $\bar{z}_t^w=(1-t)\,\mathcal{E}(\boldsymbol{X}_w)+t\,\boldsymbol{\epsilon}$, so it matches the timestep but is never updated by the editor. Another drops the evidence branch and lets the base editor restore the warped image directly. Re-noising degrades all metrics, and direct restoration keeps drag accuracy but changes the source identity and appearance, while the model-evolved evidence branch balances both.
 
\begin{table}[t]
\caption{Comparison with design alternatives.}
\label{tab:design_alternatives}
\centering
\footnotesize
\begin{tabular}{llccc}
\toprule
\textbf{Benchmark} & \textbf{Setting} & \textbf{MD$\downarrow$} & \textbf{CP$\uparrow$} & \textbf{PF$\uparrow$} \\
\midrule
\multirow{3}{*}{\textsc{DragBench-DR}}
& w/o RAR                         & 25.41 & 92.50 & 77.14 \\
& Global blending                 & 33.50 & 83.92 & 77.55 \\
& \cellcolor{recaGreen}RAR (ours) & \cellcolor{recaGreen}\textbf{20.03} & \cellcolor{recaGreen}\textbf{93.15} & \cellcolor{recaGreen}\textbf{84.82} \\
\midrule
\multirow{3}{*}{\textsc{DragBench-SR}}
& Direct re-noising               & 19.01 & 94.68 & 82.91 \\
& Direct restoration              & \textbf{18.06} & 92.28 & 81.06 \\
& \cellcolor{recaGreen}\emph{MoRe-Drag} & \cellcolor{recaGreen}18.54 & \cellcolor{recaGreen}\textbf{96.36} & \cellcolor{recaGreen}\textbf{86.36} \\
\bottomrule
\end{tabular}
\end{table}
 
\subsection{Robustness to Input Masks}\label{appendix:mask_robustness}
\emph{MoRe-Drag} takes handle--target points and a coarse user mask as input, and does not require a pixel-accurate mask. To test its sensitivity to the mask, we perturb each deformation mask of \textsc{DragBench-SR} before warping by erosion, dilation, and removing a contiguous part, each at a mild and a strong level. The warped image and all role masks are then recomputed from the perturbed mask. We also test automatic masks predicted by SAM~\cite{DBLP:conf/iccv/KirillovMRMRGXW23} with the handle points as prompts. As shown in Table~\ref{tab:mask_robustness}, mild perturbations only cause small drops. Erosion and missing regions mainly hurt MD, since part of the motion evidence is lost. Strong dilation keeps MD but lowers CP and PF, since unrelated content is also moved. SAM masks are often coarse or only cover a local part of the object, which leads to a clear drop. We therefore use the coarse user mask by default and treat automatic masks as an optional mode.
 
\begin{table}[t]
\caption{Robustness to perturbed and automatically predicted masks on \textsc{DragBench-SR}. IoU is computed with respect to the original mask.}
\label{tab:mask_robustness}
\centering
\footnotesize
\begin{tabular}{llcccc}
\toprule
\textbf{Input mask} & \textbf{Level} & \textbf{IoU} & \textbf{MD$\downarrow$} & \textbf{CP$\uparrow$} & \textbf{PF$\uparrow$} \\
\midrule
\rowcolor{recaGreen}
Original & -- & 1.000 & 18.54 & 96.36 & 86.36 \\
\midrule
\multirow{2}{*}{Erosion}        & Mild   & 0.818 & 21.11 & 95.80 & 84.64 \\
                                & Strong & 0.683 & 24.17 & 92.73 & 85.00 \\
\midrule
\multirow{2}{*}{Dilation}       & Mild   & 0.802 & 18.30 & 94.68 & 85.25 \\
                                & Strong & 0.600 & 18.45 & 92.45 & 72.92 \\
\midrule
\multirow{2}{*}{Missing region} & Mild   & 0.804 & 21.67 & 92.45 & 85.25 \\
                                & Strong & 0.620 & 30.60 & 91.33 & 82.07 \\
\midrule
SAM (point prompt)              & --     & 0.361 & 24.91 & 83.51 & 70.62 \\
\bottomrule
\end{tabular}
\end{table}

\subsection{Effect of Disocclusion Severity}\label{appendix:disocclusion}
Large disoccluded regions make the warped image less informative. We measure this with the disocclusion ratio $r_{\mathrm{dis}}=|\boldsymbol{M}_i|/|\boldsymbol{M}_r\cup\boldsymbol{M}_i|$, and split \textsc{DragBench-SR} into three groups of similar size. As shown in Table~\ref{tab:disocclusion}, the high-disocclusion group has a clearly worse MD, while CP and PF remain stable. The low and medium groups are not strictly ordered, since $r_{\mathrm{dis}}$ only partly reflects the difficulty and the edit type also matters. Thus, severe disocclusion mainly hurts spatial accuracy, which is consistent with the failure cases in Appendix~\ref{appendix:failure_cases}.
 
\begin{table}[t]
\caption{Results under different disocclusion severities on \textsc{DragBench-SR}.}
\label{tab:disocclusion}
\centering
\footnotesize
\begin{tabular}{lccccc}
\toprule
\textbf{Severity} & \textbf{$r_{\mathrm{dis}}$} & \textbf{\# Samples} & \textbf{MD$\downarrow$} & \textbf{CP$\uparrow$} & \textbf{PF$\uparrow$} \\
\midrule
Low    & $[0, 0.38]$    & 33 & 17.56 & 96.21 & 86.36 \\
Medium & $[0.38, 0.53]$ & 34 & 15.26 & 96.97 & 84.38 \\
High   & $[0.53, 0.75]$ & 33 & 23.77 & 96.88 & 89.84 \\
\bottomrule
\end{tabular}
\end{table}

\subsection{Effect of Instruction Interaction}\label{appendix:instruction_interaction_details}
\begin{figure}[t]
    \centering
    \includegraphics[width=\linewidth]{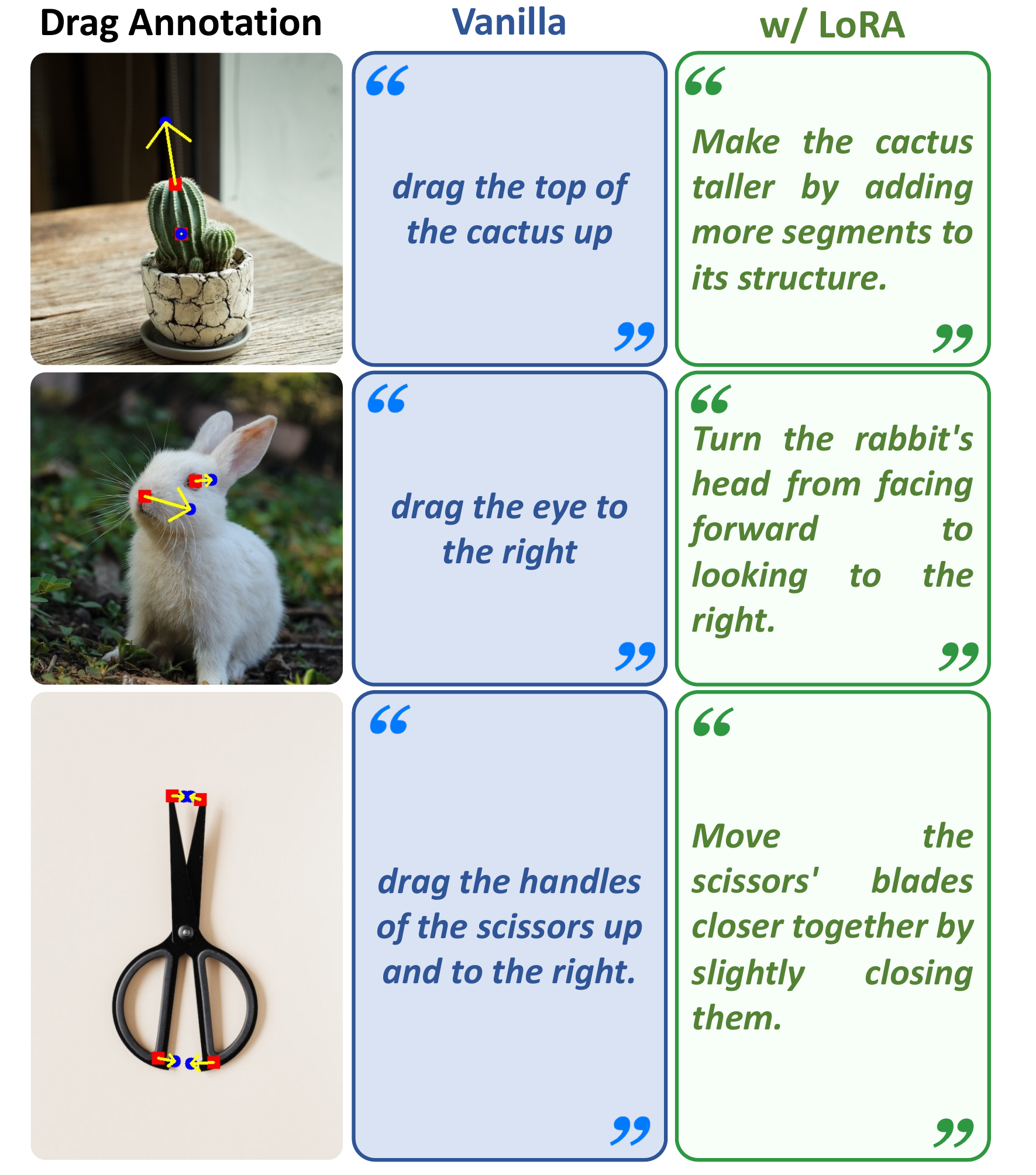}
    \caption{
    Additional comparisons of drag-aware instruction inference. 
    Each column shows a drag-annotated input, the instruction inferred by the vanilla MLLM, and the instruction inferred by the LoRA-adapted MLLM. 
    LoRA adaptation improves grounding to the visual drag annotations and produces instructions that better reflect the intended object motion or deformation.
    }
    \label{fig:instruction_inference_more}
\end{figure}

\begin{table}[t]
\centering
\caption{Quantitative analysis of instruction interaction on \textsc{DragBench-SR}.}
\label{tab:instruction_interaction_analysis}
\resizebox{0.5\linewidth}{!}{
\begin{tabular}{cccc}
\toprule
& MD$\downarrow$ & CP$\uparrow$ & PF$\uparrow$ \\
\midrule
\textbf{Vanilla}   & $19.94_{\pm0.27}$ & $95.12_{\pm0.13}$ & $78.63_{\pm1.38}$ \\
\textbf{w/ LoRA}  & ${18.54_{\pm0.22}}$  & ${96.36_{\pm0.38}}$  & ${86.36_{\pm4.06}}$ \\
\bottomrule
\end{tabular}
}
\end{table}
We provide additional examples of drag-aware instruction inference in Fig.~\ref{fig:instruction_inference_more}. Each column shows one drag-annotated input together with the instructions inferred by the vanilla MLLM and the LoRA-adapted MLLM. Compared with the vanilla model, the LoRA-adapted model more accurately identifies the edited subject and converts the drag geometry into an appropriate editing intent. Table~\ref{tab:instruction_interaction_analysis} quantifies the effect of drag-aware instruction inference on \textsc{DragBench-SR}. Compared with the vanilla MLLM, the LoRA-adapted model improves drag accuracy, reducing MD from 19.94 to 18.54, while also increasing CP and PF. These results indicate that LoRA adaptation produces instructions that are better aligned with the drag annotations and more beneficial for downstream editing.

The prompt template used for instruction inference is shown in Box~\ref{box:instruction_prompt}. During inference, the model receives the source image and its drag-annotated version, and outputs a natural-language instruction describing the intended edit. The same prompt is used for both the vanilla MLLM and the LoRA-adapted MLLM.

\paragraph{Effect of instruction detail.}
We investigate whether more detailed text improves editing performance. On 25 samples, we compare the instruction inferred by our LoRA-adapted MLLM with a detailed instruction generated by GPT-5 and a fixed restoration instruction. As shown in Table~\ref{tab:instruction_detail}, detailed instructions do not improve drag accuracy over the inferred instructions, and a fixed restoration instruction also retains comparable spatial accuracy. The inferred instructions better preserve the source concept than the fixed restoration instruction, with stronger prompt following. These results support using a short inferred instruction to guide semantic restoration, while motion evidence provides spatial control.

\begin{table}[t]
\caption{Effect of instruction detail on 25 samples from \textsc{DragBench-SR}.}
\label{tab:instruction_detail}
\centering
\footnotesize
\begin{tabular}{lccc}
\toprule
\textbf{Text condition} & \textbf{MD$\downarrow$} & \textbf{CP$\uparrow$} & \textbf{PF$\uparrow$} \\
\midrule
Auto-inferred drag-aware instruction & 17.30 & 98.00 & 85.72 \\
Detailed GPT-5-generated instruction & 17.67 & 95.00 & 86.00 \\
Fixed restoration instruction & 16.97 & 91.00 & 82.00 \\
\bottomrule
\end{tabular}
\end{table}

\begin{figure}[h]
    \centering
    \includegraphics[width=\linewidth]{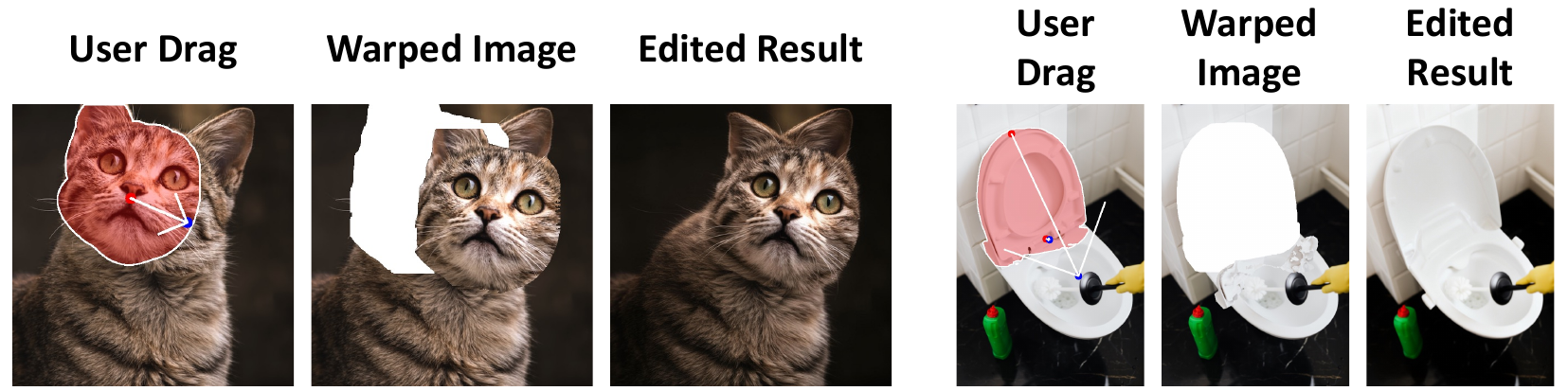}
    \caption{Failure cases. When the warped observation contains severe errors or ambiguous disocclusions, the injected motion evidence may not faithfully represent the intended edit. In these cases, \emph{MoRe-Drag} may favor plausible restoration over the desired motion, such as incomplete head manipulation or unsuccessful lid articulation.}
    \label{fig:failure_cases}
\end{figure}
\begin{figure}[h]
    \centering
    \includegraphics[width=\linewidth]{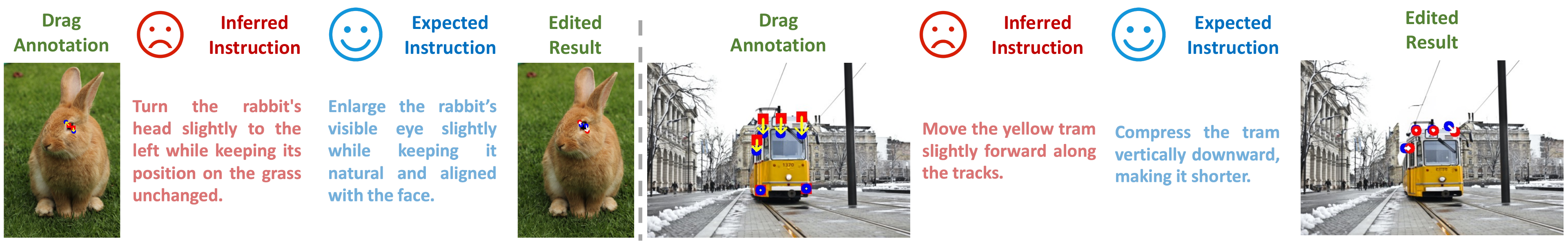}
    \caption{Failure cases of drag-aware instruction inference. The LoRA-adapted MLLM may still infer imperfect instructions when drag annotations are too close to each other, especially for fine-grained shape changes.}
    \label{fig:failure_cases_mllm}
\end{figure}

\begin{table}[!h]
\centering
\setlength{\tabcolsep}{3.2pt}
\renewcommand{\arraystretch}{0.82}
\caption{Effect of switching timestep \(t_c\) on \textsc{DragBench-SR}.}
\label{tab:t_c_analysis}

\begin{tabular}{@{}lcccccc@{}}
\toprule
Step \(k\) & 4 & 8  & 16 & 20 & 24 & 28 \\
\(t_c\) & 0.87 & 0.73 & 0.47 & 0.33 & 0.20 & 0.07 \\
\midrule
MD \(\downarrow\) & $21.01_{\pm0.1}$ & $19.27_{\pm0.25}$ & $18.59_{\pm0.32}$ & $17.98_{\pm0.09}$ & $17.52_{\pm0.24}$ & $17.36_{\pm0.16}$ \\
CP \(\uparrow\) & $96.09_{\pm0.63}$ & $97.10_{\pm0.13}$ & $97.47_{\pm0.25}$ & $97.10_{\pm0.13}$ & $95.96_{\pm0.50}$ & $96.46_{\pm0.26}$ \\
PF \(\uparrow\) & $83.33_{\pm2.68}$ & $85.10_{\pm1.99}$ & $80.08_{\pm1.29}$ & $81.06_{\pm0.51}$ & $79.31_{\pm0.74}$ & $80.94_{\pm0.13}$ \\
\bottomrule
\end{tabular}
\end{table}

\subsection{Hyperparameter Analysis}\label{appendix:hyperparameter_analysis}
We analyze two related hyperparameters on \textsc{DragBench-SR}: the number of
Region-Aware Recomposition (RAR) steps and the switching step of Generation-Stage Adaptive Conditioning (GAC). Although they are defined separately, they play a similar role in practice: increasing the RAR steps injects motion evidence for a longer interval, while delaying the GAC switch keeps the generation branch conditioned on the warped observation for more steps. Both therefore strengthen the influence of motion evidence during sampling. For this reason, we vary them synchronously and evaluate five paired settings, as shown in Table~\ref{tab:hyperparameter_analysis}.

As the RAR steps and GAC switching step increase, MD consistently decreases, indicating that stronger motion-evidence grounding improves drag precision. However, after the $(16,12)$ setting, the reduction in MD becomes marginal, while CP and PF show slight declines. This suggests that the injected motion evidence is already sufficient at $(16,12)$, and further increasing its influence brings limited gains in drag accuracy while slightly weakening semantic consistency. We therefore use $(16,12)$ as the default setting.

We further isolate the effect of the GAC switching step by fixing the number of RAR steps to 16. The quantitative results are reported in Table~\ref{tab:t_c_analysis}. When the switch occurs too early, motion evidence is removed before the spatial layout is sufficiently anchored, leading to degraded drag accuracy. Conversely, keeping the warped-observation conditioning for too long leaves insufficient sampling steps for instruction-guided refinement, often producing less natural results and prompt following. This reveals a trade-off between motion grounding and semantic refinement, and supports our choice of an intermediate switching step.

\subsection{Failure Cases}\label{appendix:failure_cases}
We present two representative failure cases in Fig.~\ref{fig:failure_cases}, where inaccurate warped observations weaken the usefulness of motion evidence. Our method is designed to incorporate motion evidence into the editing generation process, and it works reliably when the warped observation provides a reasonably consistent indication of the desired motion. However, in these cases, the motion evidence becomes ambiguous and misleading. For example, in the cat case, the drag point is insufficiently constrained for the intended rotation, which would typically require an additional anchor point; in the toilet case, the warped image does not provide a clear geometry for closing the lid. As a result, the editor tends to treat the corrupted and ambiguous regions as content to be restored rather than reliable evidence of the intended motion, leading to edits that remain visually plausible but only partially follow the target articulation. More generally, the warping in \emph{MoRe-Drag} is purely 2D. It does not know which part is in front, which part becomes hidden, or what a newly exposed surface should look like. Edits with large disoccluded regions or strong 3D motion, such as rotating an object, are therefore harder. Future work may build the motion evidence with depth and visibility information.

Another potential issue is that nearby handles with opposite displacements can partially cancel during motion interpolation, weakening the intended deformation. Region decomposition remains well defined, but cannot recover motion information lost in this upstream step.

Fig.~\ref{fig:failure_cases_mllm} shows failure cases of drag-aware instruction inference. The LoRA-adapted MLLM may infer imperfect instructions when drag annotations are spatially too close, particularly for fine-grained shape changes, since the annotations may occlude local image content and provide incomplete visual evidence. As a result, subtle shape deformation may be misinterpreted as object motion or orientation change. Nevertheless, the injected motion evidence can partially compensate for suboptimal textual guidance by directly constraining the edited region. As shown in the tram example, \emph{MoRe-Drag} still produces a motion-faithful result despite the imperfect inferred instruction. Future work may investigate alternative representations of drag information for MLLMs to further improve robustness.

\begin{figure}
    \centering
    \includegraphics[width=\linewidth]{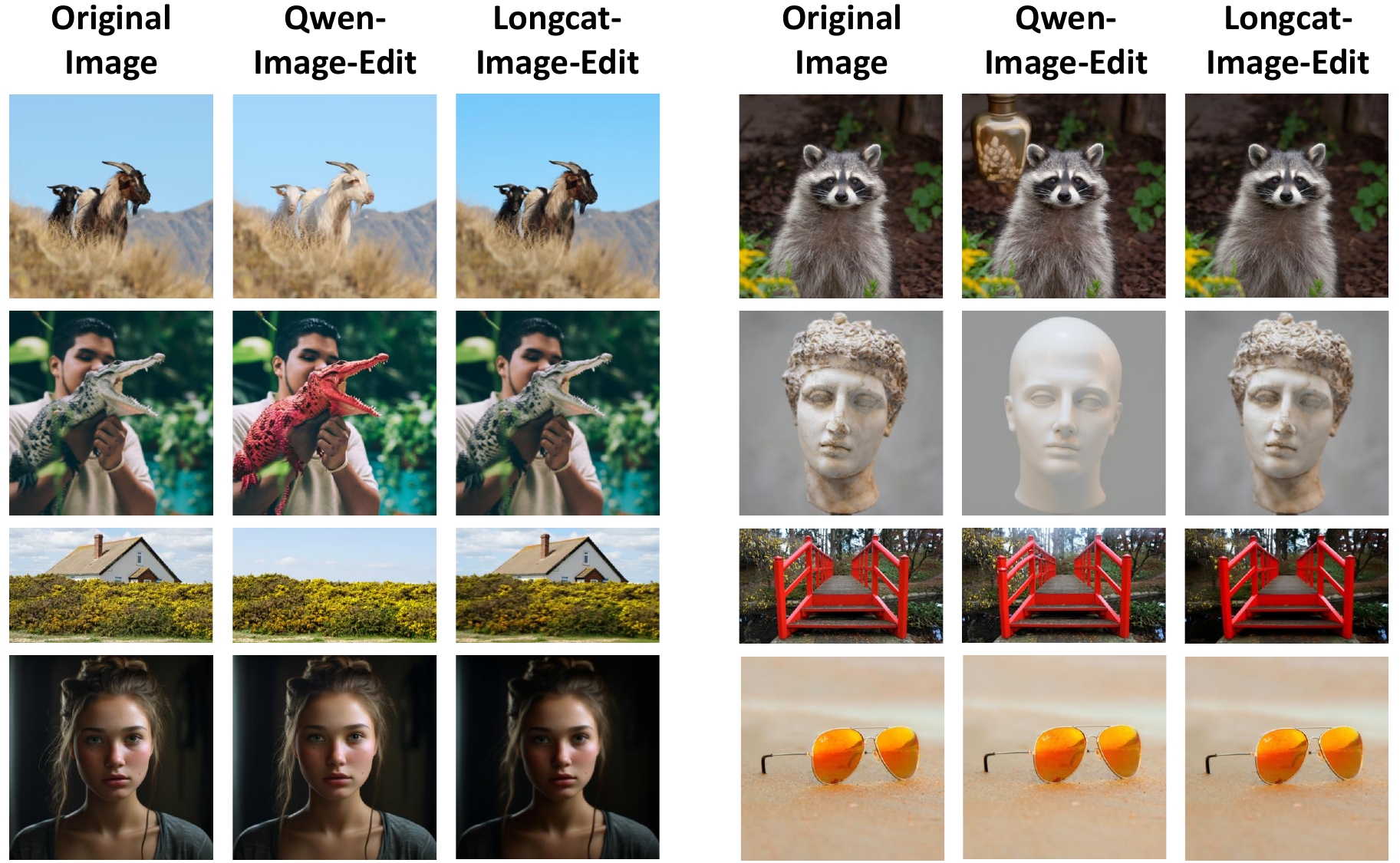}
    \caption{\textbf{Comparison of base editor reconstruction ability.} Under null conditions, LongCat-Image-Edit reconstructs the source images more faithfully than Qwen-Image-Edit, with better preservation of identity, structure, and appearance. This stronger reconstruction stability benefits motion-evidence injection and non-edited region preservation in \emph{MoRe-Drag}.}
    \label{fig:comparison_base_model}
\end{figure}
\subsection{Comparison of Base Editor Reconstruction Ability}
\label{appendix:base_model}
\emph{MoRe-Drag} relies on the base editor's null-conditioned reconstruction ability, since the injected motion evidence should guide the intended deformation while the editor is expected to faithfully preserve content outside the edited regions. We therefore compare the reconstruction behavior of different base editors under null conditions. Fig.~\ref{fig:comparison_base_model} shows qualitative examples, and Table~\ref{tab:base_reconstruction_quality} reports reconstruction LPIPS and DINO similarity together with downstream editing performance on \textsc{DragBench-SR}. LongCat-Image-Edit~\cite{DBLP:journals/corr/abs-2512-07584} reconstructs the source images more faithfully than Qwen-Image-Edit~\cite{wu2025qwenimagetechnicalreport} and achieves stronger editing results despite its smaller generator. This suggests that reconstruction stability contributes to the performance difference between the two variants.

\begin{table}[t]
\caption{Null-conditioned reconstruction quality of the two base editors and downstream editing results of their \emph{MoRe-Drag} variants on \textsc{DragBench-SR}.}
\label{tab:base_reconstruction_quality}
\centering
\footnotesize
\begin{tabular}{lccccc}
\toprule
\textbf{Base editor} & \textbf{Recon. LPIPS$\downarrow$} & \textbf{DINO Sim.$\uparrow$} & \textbf{MD$\downarrow$} & \textbf{CP$\uparrow$} & \textbf{PF$\uparrow$} \\
\midrule
Qwen-Image-Edit & 0.1533 & 0.9151 & 23.85 & 90.15 & 70.96 \\
LongCat-Image-Edit & 0.1373 & 0.9466 & 18.54 & 96.36 & 86.36 \\
\bottomrule
\end{tabular}
\end{table}

We further examine this relationship within a single editor by grouping 205 samples from \textsc{DragBench-DR} according to the DINO similarity between each source image and its null-conditioned reconstruction by LongCat-Image-Edit. Table~\ref{tab:reconstruction_quality_groups} reports the downstream editing results for each group. The low-quality group has the worst drag accuracy and concept preservation, but the high and medium groups are not consistently ordered. The low-quality group also contains few samples. Severe reconstruction errors can therefore indicate a risk of editing failure, while reconstruction similarity alone does not determine editing quality. These observations support the use of a base editor with stable reconstruction, without assuming that reconstruction quality fully predicts downstream performance.

\begin{table}[t]
\caption{Downstream editing results on 205 \textsc{DragBench-DR} samples grouped by LongCat-Image-Edit's null-conditioned reconstruction DINO similarity.}
\label{tab:reconstruction_quality_groups}
\centering
\footnotesize
\begin{tabular}{lccccc}
\toprule
\textbf{Quality} & \textbf{DINO similarity} & \textbf{\# Samples} & \textbf{MD$\downarrow$} & \textbf{CP$\uparrow$} & \textbf{PF$\uparrow$} \\
\midrule
High & $s \geq 0.90$ & 176 & 20.31 & 94.17 & 84.96 \\
Medium & $0.80 \leq s < 0.90$ & 23 & 16.68 & 89.22 & 83.11 \\
Low & $s < 0.80$ & 6 & 24.76 & 78.17 & 87.28 \\
\bottomrule
\end{tabular}
\end{table}

\subsection{Runtime and Memory Analysis}\label{appendix:complexity}
\begin{table}[t]
\centering
\caption{Runtime and memory comparison. We report the average generation time per image and
peak GPU memory over 10 samples.}
\label{tab:runtime_memory}
\resizebox{\linewidth}{!}{
\begin{tabular}{lccc}
\toprule
Method & Sampling steps & Avg. time / image (s) & Peak GPU memory (GB) \\
\midrule
Baseline editor & 30 & 30.74 & 16.05 \\
\emph{MoRe-Drag} & 30 & 41.31 & 17.18 \\
\bottomrule
\end{tabular}
}
\end{table}
We further report the computational cost of \emph{MoRe-Drag}. All measurements are conducted on
an NVIDIA RTX 4090 GPU with CPU offloading enabled. We randomly select 10 samples and report
the average generation time per image and the peak GPU memory during inference. The default
setting uses 30 sampling steps, applies region-aware latent recomposition from step 1 to step 16, and
therefore introduces an additional evidence-branch forward pass for approximately half of the
sampling trajectory.

As shown in Table~\ref{tab:runtime_memory}, \emph{MoRe-Drag} increases the average generation
time from 30.74s to 41.31s. This overhead is expected, since the evidence reconstruction branch
requires extra MMDiT forward passes during the RAR interval. Under our default setting, RAR is
applied for 16 out of 30 sampling steps, and the observed runtime increase is roughly consistent with
the additional forward computation. In contrast, the peak GPU memory only increases from 16.05 GB
to 17.18 GB. These results indicate that \emph{MoRe-Drag} improves drag
editing performance with a moderate runtime overhead and limited additional memory usage.

\section{Declaration of LLM Usage}
\label{appendix:llm_usage}
We use LLMs/MLLMs in data construction and evaluation. For data construction, we use Qwen3.5-9B~\cite{qwen3.5} to filter camera-centric samples, parse motion-centric editing instructions, and generate natural-language instructions for \textsc{DragBench-SR} and \textsc{DragBench-DR}. For semantic evaluation, we follow the LLM-based evaluation protocol of DreamBench++~\cite{DBLP:conf/iclr/PengCTQDBHG0X25} and use GPT-4o to compute the PF score and CP score. We also use GPT-5 to generate detailed instructions for the ablation on instruction detail.

\section{Prompts Used in Experiments}
\label{appendix:prompts}
\begin{promptbox}[box:instruction_prompt]{Prompt template used for drag-aware instruction inference}
\small
\textbf{System Prompt.}
You are a helpful assistant for drag-based image editing. Given (1) a source image and (2) an image overlaid with drag annotations, infer the intended edit and output a natural-language editing instruction.

\textbf{Annotation Meaning.}
A red square denotes the handle point, a blue circle denotes the target point, and a yellow arrow denotes the drag direction from the handle point to the target point.

\textbf{Output Requirement.}
Output only the editing instruction.
\end{promptbox}

\begin{promptbox}[box:drag_quality_prompt]{Prompt used for VLM-based drag-annotation quality scoring}
\small
\textbf{Task Definition.}
You are a strict judge for mined drag-annotation quality in an image-editing dataset.

You will be given three images in this exact order:
\begin{enumerate}[leftmargin=1.2em,itemsep=0pt,topsep=1pt]
    \item Source image before editing.
    \item Source image with drag guidance drawn on it.
    \item Target image after editing.
\end{enumerate}

Use the images as primary evidence.
Use the instruction as supporting evidence.
Be conservative: if the evidence is weak, ambiguous, or the drag signal is tiny and unclear, choose the lower score.

\textbf{Drag Annotation Legend.}
In image 2, the drag guidance uses this legend:
\begin{itemize}[leftmargin=1.2em,itemsep=0pt,topsep=1pt]
    \item red circle: handle/source point
    \item blue circle: target point
    \item white arrow: drag direction from handle to target
\end{itemize}

\textbf{Output Format.}
Return exactly one valid JSON object and nothing else.
Do not use markdown.
Do not wrap the JSON in code fences.
The JSON must be on a single line.
Always output all required fields.
Never output null, missing fields, placeholders, or non-integer scores.
If uncertain, still choose one valid integer score for every score field.

\textbf{Output Schema Example.}
\texttt{\{"camera\_motion\_score":3,"camera\_motion\_reason":"The framing and background stay almost identical between source and target.","instruction\_alignment\_score":2,"instruction\_alignment\_reason":"The drag arrows roughly follow the requested motion but only partially cover the edited part.","target\_alignment\_score":3,"target\_alignment\_reason":"The drag guidance clearly matches the visible local change in the target image.","confidence":"high","summary":"Reliable local drag annotation with negligible camera change."\}}

\textbf{Field Definitions.}
\begin{itemize}[leftmargin=1.2em,itemsep=0pt,topsep=1pt]
    \item \texttt{camera\_motion\_score}: integer in \(\{0,1,2,3\}\)
    \item \texttt{instruction\_alignment\_score}: integer in \(\{0,1,2,3\}\)
    \item \texttt{target\_alignment\_score}: integer in \(\{0,1,2,3\}\)
    \item \texttt{confidence}: one of \texttt{high}, \texttt{medium}, or \texttt{low}
    \item all reason and summary fields should be short and concrete
\end{itemize}

\textbf{Scoring Rubric.}

\textbf{1. Camera motion score.}
The \texttt{camera\_motion\_score} compares image 1 and image 3 only.
\begin{itemize}[leftmargin=1.2em,itemsep=0pt,topsep=1pt]
    \item \textbf{3:} no noticeable camera motion; framing, viewpoint, and background layout are essentially unchanged
    \item \textbf{2:} very slight camera change or tiny crop/scale shift, but it is barely noticeable
    \item \textbf{1:} noticeable camera/view/framing shift that is not huge but should still be penalized
    \item \textbf{0:} large camera pan, zoom, tilt, rotation, viewpoint change, or global reframing
\end{itemize}

\textbf{2. Instruction alignment score.}
The \texttt{instruction\_alignment\_score} compares the instruction and image 2.
\begin{itemize}[leftmargin=1.2em,itemsep=0pt,topsep=1pt]
    \item \textbf{3:} the drag guidance clearly and directly expresses the requested edit intent
    \item \textbf{2:} mostly aligned, but somewhat incomplete, partially ambiguous, or not fully precise
    \item \textbf{1:} weak alignment; some relevance exists, but the intent is hard to read, incomplete, noisy, or the drag magnitude is too small to clearly communicate the edit
    \item \textbf{0:} clearly mismatched with the instruction, or the drag guidance is so unclear that the intended edit cannot be identified
\end{itemize}

Important rule for \texttt{instruction\_alignment\_score}: tiny drags, nearly overlapping handle/target points, or arrows that barely indicate movement should receive a low score even if they are not obviously wrong.

\textbf{3. Target alignment score.}
The \texttt{target\_alignment\_score} compares image 2 and image 3.
\begin{itemize}[leftmargin=1.2em,itemsep=0pt,topsep=1pt]
    \item \textbf{3:} the drag guidance clearly predicts the target edit; following the arrows would plausibly transform the source toward the target
    \item \textbf{2:} mostly predictive of the target, with minor mismatch or incompleteness
    \item \textbf{1:} weak relation; some target change is explained but important parts are missing or inaccurate
    \item \textbf{0:} the drag guidance does not explain the target edit
\end{itemize}

\textbf{Additional Rules.}
\begin{itemize}[leftmargin=1.2em,itemsep=0pt,topsep=1pt]
    \item Favor local geometric motion or deformation evidence over vague semantic guesses.
    \item If the target change is dominated by global camera change, \texttt{target\_alignment\_score} should usually be low.
    \item Do not hallucinate hidden edits that are not visible in the images.
\end{itemize}
\end{promptbox}

\begin{promptbox}[box:benchmark_prompt]{Prompt used to generate natural-language editing instructions for \textsc{DragBench-SR} and \textsc{DragBench-DR}}
\small
You are an intent parser for drag-based image editing.

You are given two images of the same scene:
\begin{itemize}[leftmargin=1.2em,itemsep=0pt,topsep=1pt]
    \item \textbf{Source image}: the original image.
    \item \textbf{Annotated image}: the same image with drag annotations.
\end{itemize}

Your task is to infer the single most appropriate editing instruction implied by the drag annotations.

\textbf{Image roles.}
\begin{enumerate}[leftmargin=1.4em,itemsep=0pt,topsep=1pt]
    \item Use Source image to identify what objects, parts, people, animals, and structures truly exist in the scene.
    \item Use Annotated image only to read the drag annotations and motion geometry.
    \item Do not infer any editable subject that is not grounded in Source image.
\end{enumerate}

\textbf{Annotation meaning in Annotated image.}
\begin{itemize}[leftmargin=1.2em,itemsep=0pt,topsep=1pt]
    \item red square: handle point
    \item blue circle: target point
    \item yellow arrow: intended drag direction from handle to target
\end{itemize}

\textbf{Task objective.}
Infer what existing content should be edited, how it should be edited, and to what extent, based on the drag guidance. Return exactly one concise natural-language editing instruction.

\textbf{Edits that fit drag-based editing include:}
\begin{itemize}[leftmargin=1.2em,itemsep=0pt,topsep=1pt]
    \item moving an object or a coherent semantic part
    \item rotating, turning, tilting, or reorienting an object or part
    \item changing pose of a person, animal, or articulated object
    \item opening or closing an articulated structure
    \item extending, shortening, widening, narrowing, enlarging, shrinking, bending, stretching, compressing, or reshaping an existing part or structure
\end{itemize}

\textbf{Edits that do not fit drag-based editing include:}
\begin{itemize}[leftmargin=1.2em,itemsep=0pt,topsep=1pt]
    \item adding, removing, replacing, or duplicating objects
    \item changing identity
    \item changing background or unrelated scene content
    \item changing style, texture, material, lighting, weather, or rendering
\end{itemize}

\textbf{Inference procedure.}
\begin{enumerate}[leftmargin=1.4em,itemsep=0pt,topsep=1pt]
    \item Locate the visual content nearest each handle point in Source image.
    \item Decide the correct semantic subject of editing: a whole object, a coherent part, an articulated part, or a local structure.
    \item Prefer the most natural semantic level. If a handle lies on a small attached feature, first consider whether the drag is better explained by editing the larger connected structure rather than moving that small feature by itself.
    \item Use the handle-to-target geometry to determine the edit type. Consider whether the drag indicates translation, rotation, reorientation, pose change, articulation, scaling, or local shape deformation.
    \item Use the target location and drag distance to determine the intended final state or magnitude of the edit.
    \item If multiple drag pairs are present, combine them into one coherent edit.
    \item Output one instruction describing only the intended edit, while keeping all unrelated regions unchanged.
\end{enumerate}

\textbf{Guidelines for choosing the semantic subject.}
\begin{itemize}[leftmargin=1.2em,itemsep=0pt,topsep=1pt]
    \item Prefer a realistic, coherent edit over isolated motion of a tiny local feature.
    \item If the handle lies on a rigidly attached facial feature such as the nose, nostril, cheek, eye corner, or ear, consider whether the intended edit is actually the head, face orientation, expression, or mouth/jaw articulation.
    \item If the handle lies on a tip, contour, boundary, or elongated protruding part such as hair, cloth edge, leaf, tail, horn, ear, branch, handle, or limb, consider whether the intended edit is local extent or shape change of that part.
    \item If the handle lies on a naturally articulated structure such as a jaw, mouth, beak, eyelid, lid, claw, arm, or leg, consider whether the intended edit is opening, closing, bending, raising, lowering, or pose change.
    \item Prefer the interpretation that best matches visible structure, drag direction, target position, and natural editing semantics.
\end{itemize}

\textbf{Guidelines for determining edit type.}
\begin{itemize}[leftmargin=1.2em,itemsep=0pt,topsep=1pt]
    \item Use translation when the drag suggests relocating the subject without changing its shape or orientation.
    \item Use rotation or reorientation when the drag suggests turning the subject around a coherent base or axis.
    \item Use pose change when the drag changes the configuration of an articulated body or part.
    \item Use opening or closing when connected parts are being separated or brought together around a hinge-like structure.
    \item Use scaling when the drag changes the overall size of the subject.
    \item Use local deformation when the drag changes the length, width, height, curvature, span, or shape of an existing part.
\end{itemize}

\textbf{Guidelines for determining edit extent.}
\begin{itemize}[leftmargin=1.2em,itemsep=0pt,topsep=1pt]
    \item Infer the final intended position, angle, pose, opening state, size, or shape from the target point rather than describing only the drag direction.
    \item Choose the resulting state that best matches the drag geometry and the visible structure.
    \item Be specific about the resulting orientation or structural change when it is visually implied.
\end{itemize}

\textbf{Preservation constraints.}
\begin{itemize}[leftmargin=1.2em,itemsep=0pt,topsep=1pt]
    \item When appropriate, include a brief constraint stating which closely related parts should remain unchanged.
    \item Preserve all unrelated objects, background regions, lighting, style, and scene content.
    \item Keep the constraint realistic and local to the edit. Mention unchanged parts only when they help specify the intended edit more precisely.
    \item Prefer preserving the surrounding structure that naturally anchors the edited subject, such as the rest of the body, the rest of the face, nearby connected parts, or the object's main base.
    \item Do not add unnecessary preservation details. Only mention the most relevant unchanged parts.
    \item The instruction may include a short clause such as while keeping the rest of the subject unchanged, when that makes the intent clearer.
\end{itemize}

\textbf{Output requirements.}
\begin{enumerate}[leftmargin=1.4em,itemsep=0pt,topsep=1pt]
    \item Return exactly one concise edit instruction.
    \item Output only the instruction and nothing else.
    \item Do not output explanations, reasoning steps, JSON, or alternatives.
    \item Do not mention annotations, colors, points, arrows, coordinates, Image 1, or Image 2.
    \item The instruction may include a brief preservation constraint about what should remain unchanged.
    \item Keep unrelated content unchanged.
\end{enumerate}

\textbf{Examples of good outputs.}
\begin{itemize}[leftmargin=1.2em,itemsep=0pt,topsep=1pt]
    \item Turn the dog's head toward the camera while keeping the body pose unchanged
    \item Open the lion's mouth while keeping the head position unchanged
    \item Raise the man's arm while keeping the rest of his pose unchanged
    \item Pull the hat brim down while keeping the hat crown in place
    \item Lengthen the table while keeping its position unchanged
    \item Widen the cup while keeping its height unchanged
    \item Extend the leaf into the gap while keeping the rest of the plant unchanged
    \item Move the fire hydrant to the left while keeping the background unchanged
\end{itemize}

\textbf{Examples of bad outputs.}
\begin{itemize}[leftmargin=1.2em,itemsep=0pt,topsep=1pt]
    \item Move the nose to the left
    \item Move one eye upward
    \item Add another leaf
    \item Make the image brighter
    \item Change the background
\end{itemize}
\end{promptbox}

\begin{promptbox}[box:pf_prompt]{Prompt used for adapted PF evaluation in the editing setting.}
\small
\textbf{Task Definition.}
You will be provided with a source image, an edited image, and an edit instruction.
As an experienced evaluator, your task is to evaluate whether the edited image follows the edit instruction while preserving the unchanged visual content from the source image, according to the scoring criteria.

\textbf{Scoring Criteria.}
When assessing the consistency of an edited image with an edit instruction, compare the source image and the edited image carefully. This evaluation can be based on several key aspects:
\begin{enumerate}[leftmargin=1.2em,itemsep=0pt,topsep=1pt]
    \item \textbf{Instruction alignment:} Determine whether the visible change in the edited image directly follows the requested edit instruction.
    \item \textbf{Edit locality and preservation:} Evaluate whether areas, subjects, identities, background, style, and visual details that are not requested to change remain consistent with the source image.
    \item \textbf{Accuracy:} Examine whether the edited target area, motion, deformation, position, appearance, or interaction is depicted correctly according to the instruction.
    \item \textbf{Completeness:} Evaluate whether all critical parts of the instruction are reflected in the edited image without omitting important requested changes.
\end{enumerate}

\textbf{Scoring Range.}
Based on these criteria, a specific integer score from 0 to 4 can be assigned to determine the level of semantic consistency:
\begin{itemize}[leftmargin=1.2em,itemsep=0pt,topsep=1pt]
    \item \textbf{Very Poor (0):} No meaningful alignment. The edited image does not follow the instruction, or the edit severely damages unrelated content.
    \item \textbf{Poor (1):} Weak alignment. The edited image only vaguely relates to the instruction and has major errors, missing changes, or unnecessary alterations.
    \item \textbf{Fair (2):} Moderate alignment. The edited image partially follows the instruction but has noticeable inaccuracies, incomplete edits, or preservation issues.
    \item \textbf{Good (3):} Strong alignment. The edited image follows most of the instruction with only minor inaccuracies or minor unintended changes.
    \item \textbf{Excellent (4):} Near-perfect alignment. The edited image faithfully executes the instruction and preserves all unrelated visual content.
\end{itemize}

\textbf{Input format.}
Every time you will receive an edit instruction and two images. The first image is the source image before editing, and the second image is the edited image after applying the instruction.

Please carefully review the source image, edited image, and edit instruction. Before giving a score, please provide a brief analysis of the above evaluation criteria, which should be very concise and accurate.

\textbf{Output Format.}

Analysis: [Your analysis]\\
Score: [Your Score]
\end{promptbox}
\begin{promptbox}[box:fixed_prompt]{Fixed prompt used in Generation-stage adaptive conditioning.}
\small
Repair this roughly warped edited image by filling all white blank holes, missing regions, and transparent-looking gaps with semantically appropriate content inferred from the surrounding image.

Do not simply fill the holes with background; reconstruct the correct object or scene content based on context, such as animal body parts, fur, grass, or other relevant structures.

Refine and correct any unnatural, distorted, warped, stretched, blurry, duplicated, or structurally inconsistent areas introduced by the coarse warp editing, especially inside and around the edited regions.

Preserve the original identity, appearance, composition, pose, background, colors, lighting, texture, perspective, and fine details as much as possible outside the damaged or edited areas.

Ensure the final image is realistic, seamless, and visually coherent, with natural transitions, clean boundaries, and no visible artifacts or inconsistencies.
\end{promptbox}
\FloatBarrier

\end{document}